\documentclass[10pt,twocolumn,letterpaper]{article}

\usepackage[pagenumbers]{iccv} 
\usepackage{booktabs,multirow,makecell,multicol}
\newcommand*{\blarrow}{\rotatebox[origin=c]{180}{$\Lsh$}}
\newcommand*{\y}{\checkmark}
\newcommand*{\n}{\textcolor{gray}{--}}

\newcommand{\fnseg}[0]{$\mathrm{FN}_\mathrm{seg}$}
\newcommand{\fncls}[0]{$\mathrm{FN}_\mathrm{cls}$}
\newcommand{\tp}[0]{$\mathrm{TP}$}

\usepackage{dblfloatfix}

\definecolor{iccvblue}{rgb}{0.21,0.49,0.74}
\definecolor{dashedblue}{HTML}{4285F4}
\usepackage[pagebackref,breaklinks,colorlinks,allcolors=iccvblue]{hyperref}
\usepackage{amsmath}
\usepackage{stmaryrd}
\usepackage[most]{tcolorbox}
\def\paperID{28} 
\def\confName{ICCV}
\def\confYear{2025}

\newcommand{\withvoid}[1]{\mathbf{#1}}
\newcommand{\withoutvoid}[1]{\mathbf{#1'}}

\newcounter{researchfinding}

\newcommand{\researchfinding}[2][]{%
  \refstepcounter{researchfinding}%
  \begin{tcolorbox}[
    enhanced,
    colback=blue!5,               
    colframe=blue!70!black,       
    boxrule=0.8pt,                
    left=1pt,right=1pt,top=0.5pt,bottom=0.5pt,
    boxsep=3pt,
  ]
  \normalsize #2
  \end{tcolorbox}
  \ifx\\#1\\\else\label{rf:#1}\fi
}

\title{What Holds Back Open-Vocabulary Segmentation?}

\author{%
Josip Šarić\textsuperscript{1}\footnotemark[1]
\qquad
Ivan Martinović\textsuperscript{2}\footnotemark[1]
\qquad
Matej Kristan\textsuperscript{1} 
\qquad
Siniša Šegvić\textsuperscript{2 \textdagger}\\
    {\normalsize \begin{tabular}{cc}
         {\textsuperscript{1}Faculty of Computer and Information Science} & 
         {\textsuperscript{2}Faculty of Electrical Engineering and Computing} \\
         University of Ljubljana, {\tt\small name.surname@fri.uni-lj.si} & University of Zagreb, {\tt\small name.surname@fer.hr} 
    \end{tabular}}
}
\begin{document}
\twocolumn[{%
  \maketitle
  \renewcommand\twocolumn[1][]{#1}%
    \captionsetup{type=figure}
    \centering
    \includegraphics[width=\textwidth]{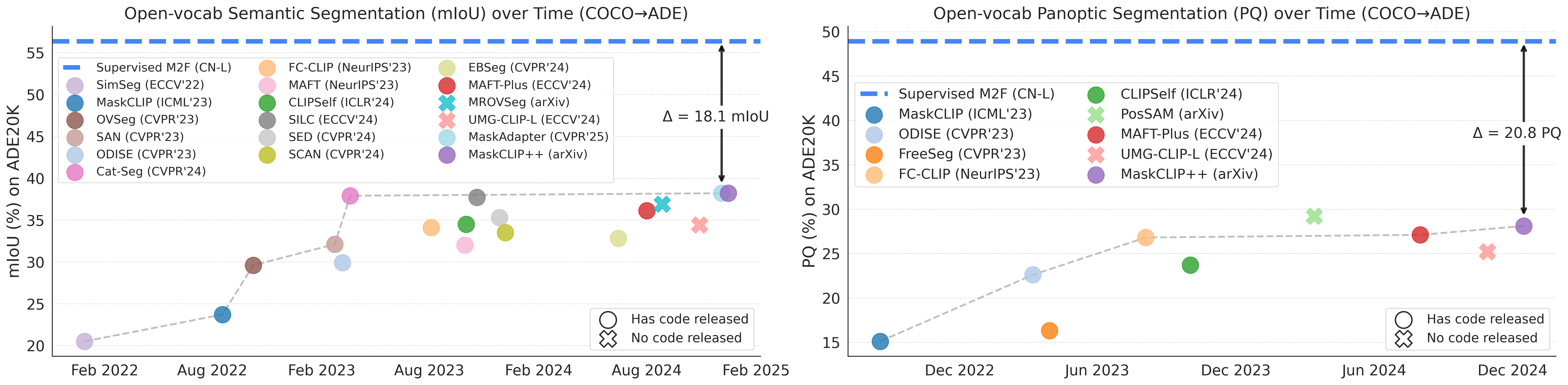}
    \caption{
        Performance over time of open-vocabulary semantic (left) and panoptic (right) segmentation methods (COCO$\rightarrow$ADE20K). The \textcolor{dashedblue}{blue dashed} line denotes 
        the Mask2Former supervised performance with a comparable backbone. Colored circles/crosses denote different open-vocabulary methods. We observe that the open-vocabulary methods have hit a plateau, lagging behind supervised models. 
    }
    \label{fig:teaser}
    \vspace{10pt}
}]

\renewcommand{\thefootnote}{\fnsymbol{footnote}}
\footnotetext[1]{Equal contribution.}

\begin{abstract}
Standard segmentation setups
are unable to deliver models
that can recognize concepts
outside the training taxonomy.
Open-vocabulary approaches
promise to close this gap
through language-image pretraining
on billions of image-caption pairs.
Unfortunately, we observe that the promise
is not delivered due to several bottlenecks
that have caused 
the performance
to plateau for almost two years.
This paper proposes novel oracle components 
that identify and decouple these bottlenecks 
by taking advantage of the groundtruth information. 
The presented validation experiments 
deliver important empirical findings
that provide a deeper insight 
into the failures of open-vocabulary models
and suggest prominent approaches 
to unlock the future 
research.
\end{abstract}    
\section{Introduction}
\label{sec:intro}
Image segmentation is 
an important task 
in computer vision, 
supporting wide 
range of applications
such as autonomous driving~\cite{cordts16cvpr},
medical imaging~\cite{ronneberger15miccai},
and remote sensing~\cite{demir18cvprw}.
The task has 
been well-studied
under different paradigms
such as semantic~\cite{long15cvpr},
instance~\cite{he17iccv}, 
and panoptic segmentation~\cite{kirillov19cvpr},
which unifies the former two.
Despite strong progress
in validation accuracy~\cite{cheng22cvpr,wang23cvpr}, 
most of the conventional methods
remain constrained to reasoning
within the training taxonomy.
This rigidity hinders
generalization capabilities
and application potential
in the wild.

Recently, vision-language models 
(VLM) 
have emerged as 
a promising solution 
for recognition beyond the training taxonomy~\cite{radford21icml,jia21icml}. 
Recognition based on 
image-text similarities
enables effortless 
vocabulary expansion at test time
by simply introducing novel
class descriptions.
Extending this capability 
to dense prediction has 
become a prominent focus 
in recent research, 
giving rise to the task 
of open-vocabulary segmentation~\cite{ghiasi22eccv,liang23cvpr}.

While early approaches 
showed promise, 
recent open-vocabulary segmentation models 
still lag significantly behind 
their in-domain counterparts. 
\Cref{fig:teaser} illustrates this gap 
for semantic (left) 
and panoptic segmentation (right), 
showing the performance 
of open-vocabulary models trained on COCO~\cite{lin14eccv} 
and evaluated on ADE20K~\cite{zhou17cvpr}, 
alongside the closed-set Mask2Former~\cite{cheng22cvpr} 
trained directly on ADE20K (\textcolor{dashedblue}{blue dashed line}). 
Notably, the best models 
trail the closed-set 
in-domain baseline 
by nearly 20 points.
Even more concerning 
is the apparent stagnation 
in open-vocabulary performance, 
despite the task’s relatively 
recent emergence 
and the high annotation costs 
of current training pipelines.
In addition to relying 
on millions of image-caption pairs, 
these methods train 
on more than 100,000 densely 
annotated images from COCO. 
In contrast, 
our experiments demonstrate 
that in-domain models
achieve comparable performance 
with as few as 300 labeled 
images from ADE20K (\cf~\cref{fig:supervised_comparison}).
This suggests 
that current open-vocabulary 
methods and setups share 
some underlying limitations
that prevent them
from closing the gap.

In this paper, 
we investigate the root causes 
of the exposed issue
in the context of
mask-transformer-based~\cite{cheng22cvpr,yu22eccv,li23cvpr,wang23cvpr}
open-vocabulary segmentation methods~\cite{yu23neurips,jiao24eccv}.
Our detailed analysis 
of top-performing methods
identifies key bottlenecks 
and uncovers several
insightful findings. 
We show that none
of the core components
in current open-vocabulary
methods are sufficiently effective:
vision-language models struggle 
with region-level classification,
and mask proposal generators 
often fail to provide 
adequate segmentation. 
In fact, many valid masks 
are produced internally, 
but discarded during inference 
due to conflicting training 
and testing objectives.
These findings raise concerns 
about the current training setup, 
where evaluation goals 
are often infeasible 
given the supervision provided. 
Finally, we outline research
directions to address
these issues and
propel open-vocabulary
segmentation forward.

\section{Related Work}
\label{sec:relatedwork}
\noindent \textbf{Image segmentation.} 
Early deep learning approaches 
treated semantic 
and instance segmentation 
as separate tasks. 
Seminal work 
in semantic segmentation
adapted convolutional 
classification networks
for dense prediction~\cite{long15cvpr}.
Subsequent work enhanced spatial details 
through atrous convolutions~\cite{chen17pami}, pyramidal pooling~\cite{zhao17cvpr}, 
and ladder-style decoders~\cite{ronneberger15miccai,kreso2017ladder,chen2018encoder}. 
On the other hand,
instance segmentation 
evolved from 
object detection pipelines~\cite{girshick14cvpr,girshick15iccv,ren15neurips}, 
adding segmentation heads 
to produce 
per-instance masks~\cite{hariharan14eccv,he17iccv}. 
Later, panoptic segmentation~\cite{kirillov19cvpr}
unified both tasks.
Most prominent 
panoptic methods
build on the 
mask transformer framework 
\cite{wang21cvpr,cheng21neurips,cheng22cvpr},
which represents a unified architecture
for all segmentation tasks.

\noindent \textbf{Vision-language models.} Contrastively pretrained vision–language models such as CLIP~\cite{radford21icml} and ALIGN~\cite{jia21icml} are central to open-vocabulary segmentation. Trained on large-scale data consisted of image–text pairs~\cite{radford21icml,schuhmann22neurips,chen2022pali}, they learn to embed both modalities in a shared semantic space, enabling direct cross-modal comparison. 
Subsequent methods enhance CLIP in three key ways: they introduce optimized classification~\cite{zhai2023sigmoid} and spatial-aware localization losses~\cite{maninis2025tips,tschannen2025siglip}; refine the training procedure~\cite{zhai2022lit,li2023inverse,li2023scaling,sun2023eva}; and curate higher-quality pre-training data~\cite{xu2023demystifying,fang2024data}.
OpenCLIP~\cite{cherti23cvpr} trains CLIP from scratch on the public LAION-5B dataset~\cite{schuhmann22neurips} and introduces ConvNeXt~\cite{liu2022convnet} 
backbones alongside standard ViTs~\cite{dosovitskiy2020image}.

\noindent \textbf{Open-vocabulary segmentation.} There are two main paradigms for open-vocabulary segmentation: (i) training-free~\cite{zhou22eccv, li2023clipsurgery, wang2024sclip, lan2024clearclip, karazija2024diffusion} and (ii) training-based~\cite{ding2022decoupling, liang2023open, xu2023side, xu2022groupvit,mukhoti2023open}. Training-free methods mostly rely on vision-language models (such as CLIP) to make zero-shot predictions. 
We focus on training-based approaches
as they offer better performance
and enable instance-level recognition.
These approaches are further divided into (i) weakly supervised~\cite{xu2022groupvit,mukhoti2023open,cha2023learning,xu2023cvprlearningfromnls} and (ii) fully supervised~\cite{ding2022decoupling, liang2023open, xu2023side, xu2023open, yu23neurips, han2023deop, xie2024cvpr, cho2024cat}, with the latter being more common. We focus on fully supervised methods, which train on COCO~\cite{lin14eccv} and evaluate on a broad suite of test benchmarks~\cite{everingham2010pascal,cordts16cvpr,zhou17cvpr,neuhold2017mapillary}. Fully supervised open-vocab methods aim to learn generic \textit{objectness} from ground-truth annotations while remaining robust to the domain shift encountered at test-time. Such methods fall into two groups according to the underlying segmentation model: (i) pixel/patch-based~\cite{cho2024cat} and (ii) mask-based~\cite{ding2022decoupling,liang2023open,xu2023open,yu23neurips,jiao24eccv,li2025mask}. CAT-Seg~\cite{cho2024cat} is the most prominent pixel-based method; it refines pixel-level cosine similarities between CLIP image and text embeddings through a cost-aggregation framework. Several subsequent works~\cite{wu2023clipself,naeem2024silc,tschannen2025siglip} evaluate their vision–language pretrained models within this CAT-Seg framework. 
Mask-based approaches classify complete masks rather than individual pixels or patches. Classification can operate in image space via mask cropping~\cite{ding2022decoupling,liang2023open} or in feature space through mask pooling~\cite{xu2023open,yu23neurips,jia21icml,li2025mask,martinovic2025dearli}. 
We focus on mask-based models, which represent the most general framework as they support semantic, instance, and panoptic segmentation.
Specifically, we study two recent unified models, FC-CLIP~\cite{yu23neurips} and MAFT+\cite{jiao24eccv}, described in \Cref{sec:preliminary}.

\section{Preliminary}
\label{sec:preliminary}
\noindent \textbf{Task definition}. Open-vocabulary segmentation
aims to partition an input image $\mathbf{I}\in\mathbb{R}^{H\times W\times 3}$
into a set of binary masks
with corresponding semantic labels:
\begin{equation}
\bigl\{(\mathbf{M}_i, c_i)\bigr\}_{i=1}^{N},\qquad \mathbf{M}_i\in\{0,1\}^{H\times W}.
\end{equation}
We denote by $N$ the number of ground-truth segments in $\mathbf{I}$,
and each segment is associated with a semantic label $c_i$.
An open-vocabulary segmentation model
is trained on a label set $\mathcal{C}_{\text{train}}$,
whereas at inference time it may
encounter
previously
unseen categories drawn from a test taxonomy $\mathcal{C}_{\text{test}}$ (\ie  $\mathcal{C}_{\text{train}}\neq\mathcal{C}_{\text{test}}$). 
We can split $\mathcal{C}_{\text{test}}$ into seen $\mathcal{C}_{\text{seen}}=\mathcal{C}_{\text{test}}\cap\mathcal{C}_{\text{train}}$ and unseen semantic categories $\mathcal{C}_{\text{unseen}}=\mathcal{C}_{\text{test}}\setminus\mathcal{C}_{\text{train}}$.
We focus on standard setup with COCO as training ($\mathcal{C}_\text{train}$) and ADE20K as evaluation dataset ($\mathcal{C}_\text{test}$).
A common approach in open-vocab segmentation encodes each class label as a CLIP-based text embedding and matches these embeddings against pixel- or region-level visual features.
\begin{figure*}[t!]
    \centering
    \includegraphics[width=\linewidth]{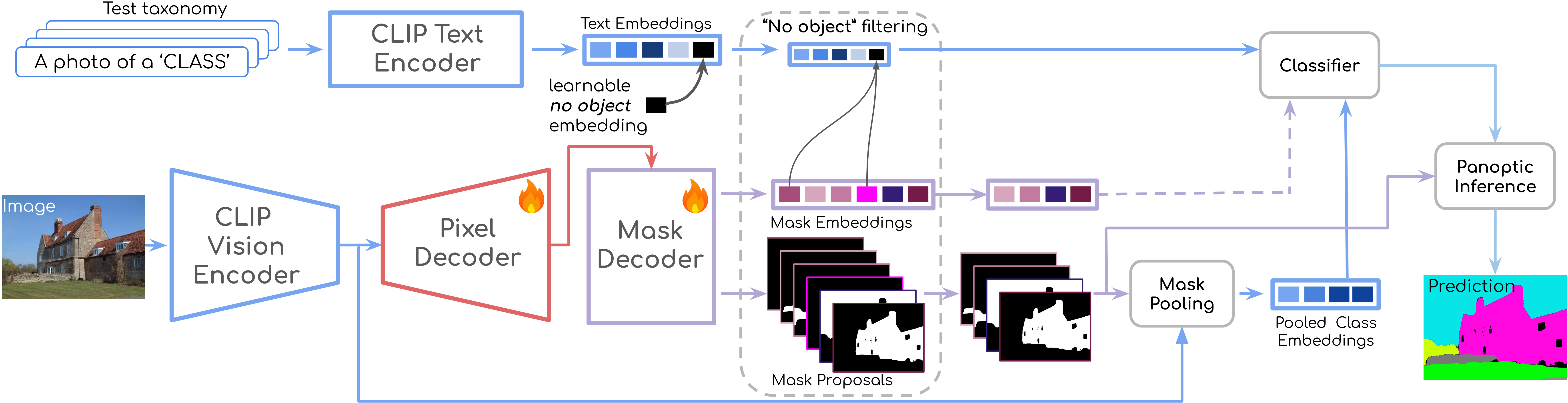}
    \caption{
    Standard mask-based pipeline for open-vocabulary segmentation, as proposed in FC-CLIP~\cite{yu23neurips} and MAFT+~\cite{jiao24eccv}. At test time, CLIP vision–text encoders supply aligned visual and textual features. A mask decoder, starting from $N$ learnable embeddings, cross-attends to the visual features to produce class-agnostic mask proposals. Masks tagged as \textit{no object} are discarded; the rest are labeled by ensembling a learned head (FC-CLIP) with a mask-pooled CLIP head. A panoptic inference then fuses masks and logits into the prediction.}
    \label{fig:overview}
\end{figure*}

\noindent \textbf{Performance metrics}. 
Our analysis emphasizes panoptic segmentation as the most comprehensive segmentation task. Hence, our primary metric is panoptic quality~\cite{kirillov19cvpr}: 
\begin{equation}
\label{eq:panoptic_quality}
\small{\mathrm{PQ}
=\underbrace{\frac{\displaystyle\sum_{(p,g)\,\in\,\mathrm{TP}}\!\mathrm{IoU}(p,g)}{|\mathrm{TP}|}}_{\text{segmentation quality (SQ)}}
\times
\underbrace{\frac{|\mathrm{TP}|}{|\mathrm{TP}|+\tfrac12|\mathrm{FP}|+\tfrac12|\mathrm{FN}|}}_{\text{recognition quality (RQ)}}}.
\end{equation}

Sets $\mathrm{TP}$, $\mathrm{FP}$, and $\mathrm{FN}$ denote the correctly matched prediction–ground truth mask-segment pairs (true positives), unmatched masks (false positives), and unmatched ground-truth segments (false negatives), respectively. 
A predicted mask $p$ matches a ground-truth segment $g$
if they share the semantic class ($c_p = c_g$)
and overlap with $\mathrm{IoU}(p,g) > 0.5$.
We compute PQ (\cf~Eq.~\ref{eq:panoptic_quality}) independently for each class and report the mean across classes.
We also report PQ\textsubscript{seen} 
and PQ\textsubscript{unseen}
as mean PQ over
the corresponding 
class subsets.

\noindent \textbf{Open-vocab methods: FC-CLIP~\cite{yu23neurips} and MAFT+~\cite{jiao24eccv}}. 
Our analysis focuses on two 
prominent recent
open-vocabulary methods:
FC-CLIP~\cite{yu23neurips} 
and MAFT+~\cite{jiao24eccv}.
Both approaches follow the dominant paradigm
in open-vocab segmentation and
pair CLIP~\cite{radford21icml}
with a mask-transformer~\cite{cheng22cvpr}.
Therefore, insights gained 
in this study generalize
to other approaches as well.
The architecture consists of three main components:
i) a vision encoder that extracts features,
ii) a pixel decoder that upsamples these features and 
iii) a mask decoder that generates
mask proposals and mask embeddings (\cf~\cref{fig:overview}).
The mask decoder decouples
segmentation into two distinct tasks:
i) mask localization and ii) mask recognition.
This design enables seamless integration
of CLIP-based recognition, 
supporting the test-time goal of identifying unseen categories $\mathcal{C}_{\text{unseen}}$ in open-vocabulary segmentation.

To this end, FC-CLIP and MAFT+
express mask-wide classification
using the similarity between
mask-pooled visual CLIP features
and textual CLIP embeddings of 
the target classes.
This setup can recognize previously
unseen categories,
provided their textual descriptions
are available at test time.
FC-CLIP~\cite{yu23neurips} preserves
vision–language alignment by freezing the CLIP
encoders during training.
In contrast, MAFT+~\cite{jiao24eccv} fine-tunes
the CLIP vision encoder but encourages
the fine-tuned features to remain
close to the pre-trained ones
by introducing an additional loss term.

We next outline components
of FC-CLIP and MAFT+ most
relevant to our analysis (\cf~\cref{fig:overview}).
Let $\mathcal{V}_{\text{CLIP}}$ and $\mathcal{T}_{\text{CLIP}}$
denote the CLIP vision and text encoders.
$\mathcal{V}_{\text{CLIP}}$ extracts features $\mathbf{F} \in \mathbb{R}^{H' \times W' \times D}$ from the input image $\mathbf{I} \in \mathbb{R}^{H \times W \times 3}$.
The mask decoder starts with $N$ learnable embeddings, attends them to the upsampled visual features, and produces mask embeddings $\mathcal{E}_{\text{mask}} \in \mathbb{R}^{N \times D}$ and pixel-to-mask scores $\mathcal{E}_{\text{pixel}} \in \mathbb{R}^{N \times H \times W}$.
Applying a sigmoid to $\mathcal{E}_{\text{pixel}}$ yields the localization maps $\boldsymbol{\sigma} \in \mathbb{R}^{N \times H \times W}$.
$\mathcal{T}_{\text{CLIP}}$ 
generates text embeddings 
$\mathbf{E}_t \in \mathbb{R}^{|\mathcal{C}| \times D}$ 
by encoding a prompt such as
\texttt{"a photo of a [class]"} 
for each class in $\mathcal{C}$.
The textual embeddings $\mathbf{E}_t$ 
act as a handcrafted linear projection,
that replaces the free weights in the standard M2F classifier.
Note that an additional $(|C| + 1)$-th 
learnable \textit{no-object} 
embedding $\mathbf{e}_{\varnothing}$ 
is appended to $\mathbf{E}_t$,
forming the extended embedding matrix $\withvoid{E} = [\mathbf{E}_t; \mathbf{e}_{\varnothing}] \in \mathbb{R}^{|C|+1 \times D}$. 
Mask-wide classification probabilities $\withvoid{P} \in \mathbb{R}^{N \times (C+1)}$ are then
obtained as
$\mathcal{E}_{\text{mask}} \cdot \withvoid{E}^\top$ followed by row-wise \emph{softmax}.
All masks classified as \textit{no-object} are discarded,
which leaves $N'$ class posteriors $\withoutvoid{P} \in \mathbb{R}^{N' \times C}$
and the corresponding localization maps $\withoutvoid{\boldsymbol{\sigma}} \in \mathbb{R}^{N' \times H \times W}$.
Note that $\mathbf{e}_{\varnothing}$
is learned on the training set.
This fact raises concerns about 
the true \textit{openness} of these methods, 
and will be one of the main focuses of our analysis.

Another key component shared by
FC-CLIP and MAFT+ 
is the mask-pooling operator,
which acts on the dense CLIP features
$\mathbf{F} \in \mathbb{R}^{H' \times W' \times D}$.
Thresholding the localization maps $\withoutvoid{\boldsymbol{\sigma}}$ yields binarized masks $\mathbf{M} \in \{0,1\}^{N' \times H \times W}$, with $\mathbf{M}_i = \llbracket \boldsymbol{\sigma}'_i \ge 0.5 \rrbracket$.
Given dense CLIP features $\mathbf{F}$ and a binary mask $\mathbf{M}_i$, mask pooling $\mathcal{MP}$ produces a mask-aggregated visual embedding $\mathbf{e}_{v_i} \in \mathbb{R}^D$:
\begin{equation}
	\label{eq:mask-pooling}
	\mathbf{e}_{v_i}=\mathcal{MP}\left(\mathbf{F}, \mathbf{M}_i\right) = \frac{\sum_{r,c}^{HW}\mathbf{F}[r,c,:]\cdot \mathbf{M}_i[r,c]}{\sum_{r,c}^{HW} \mathbf{M}_i[{r,c}]}.
\end{equation}
For each of the $N'$ predicted masks, CLIP-based probabilities can be computed by applying softmax (w/ the temperature $\tau$) to the cosine similarities between visual embedding $\mathbf{e}_{v_i} \in \mathbb{R}^{D}$ and textual embeddings $\mathbf{E}_t$:
\begin{equation}\label{eqn:zero_shot}
    \mathbf{p}_{\mathrm{CLIP}}^i = \mathrm{softmax}([{\mathbf{e}_{v_i}^T\mathbf{e}_{t_1}}, {\mathbf{e}_{v_i}^T \mathbf{e}_{t_2}}, ..., {\mathbf{e}_{v_i}^T \mathbf{e}_{t_{|\mathcal{C}|}}}], {\tau}).
\end{equation}
As the distribution $\mathbf{P'}$
is based on training categories,
open-vocab methods must incorporate
the CLIP-based distribution
$\mathbf{P}_{\text{CLIP}} \in \mathbb{R}^{N' \times D}$
during inference 
to recognize $\mathcal{C}_{\text{unseen}}$.
FC-CLIP distinguishes $\mathcal{C}_{\text{seen}}$
from $\mathcal{C}_{\text{unseen}}$
and ensembles the in-vocabulary distribution
$\mathbf{P}'$ 
with the out-vocabulary distribution $\mathbf{P}_{\text{CLIP}}$.
In contrast, MAFT+ inference
relies solely on
$\mathbf{P}_{\text{CLIP}}$,
yet obtained from the 
fine-tuned CLIP 
vision encoder.
Because mask pooling can
bottleneck unseen-class recognition capabilities,
our analysis studies the upper bounds of this operation.
\section{Empirical Analysis and Findings}
\label{sec:discussion}
Panoptic segmentation 
requires both accurate segmentation 
and correct classification 
to count a segment as
a true positive. 
This requirement 
aligns naturally 
with mask transformers, 
where mask proposal generation (segmentation) 
is decoupled from the recognition (classification). 
This led us to ask: 
\textit{how these two subtasks affect open-voc performance by themselves?} 
To investigate, we conduct
a series of experiments 
in which either 
the segmentation or classification 
gets replaced with an oracle 
- an ideal component that 
performs inference using 
ground truth information. 
These oracle-based experiments 
quantify the upper bounds 
of current open-voc methods, 
revealing how far we could push the performance 
with perfect segmentation or classification.
\subsection{Segmentation Oracle}
\begin{figure}
    \centering
    \includegraphics[width=\linewidth]{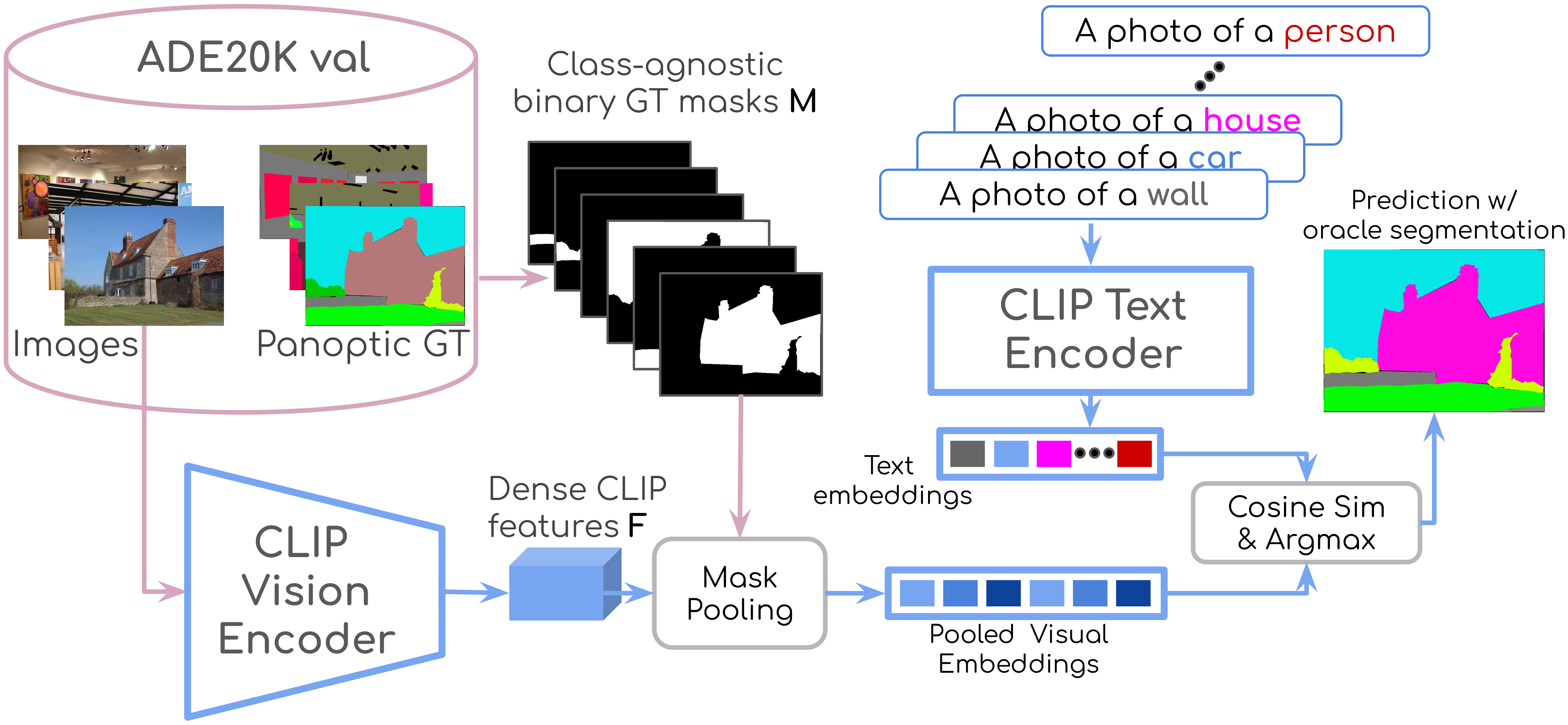}
    \caption{We estimate CLIP’s out-vocab recognition ceiling using oracle masks. For each ADE20K validation image, we extract CLIP features, pool them within the ground-truth class-agnostic masks to obtain one embedding per mask, and compute cosine similarities to the CLIP text embeddings of all ADE20K classes.}
    \label{fig:mask_pooling_oracle}
\end{figure}

\begin{table*}[h]
  \centering
  \footnotesize
  \begin{tabular}{lll   ccc   ccc   ccc}
    \toprule
      \multirow{2}{*}{Model}
      & \multirow{2}{*}{Architecture}
      & \multirow{2}{*}{\makecell{Pre-training \\ Dataset}}
      & \multicolumn{3}{c}{\text{512×512}}
      & \multicolumn{3}{c}{\text{640×640}}
      & \multicolumn{3}{c}{\text{800×1333}} \\
    \cmidrule(lr){4-6} \cmidrule(lr){7-9} \cmidrule(lr){10-12}
      & &
      & PQ$_\text{all}$ & PQ$_\text{seen}$ & PQ$_\text{unseen}$
      & PQ$_\text{all}$ & PQ$_\text{seen}$ & PQ$_\text{unseen}$
      & PQ$_\text{all}$ & PQ$_\text{seen}$ & PQ$_\text{unseen}$ \\
    \midrule
    CLIP~\cite{radford21icml}  & ViT-B-16 @ 224
        & WIT~\cite{radford21icml} & 27.0 & 36.2 & 20.2 
        & 26.3 & 35.4 & 19.5 
        & 19.6 & 28.7 & 12.8 \\
    OpenCLIP~\cite{cherti23cvpr}   & ViT-B-16 @ 224
        & LAION-2B~\cite{schuhmann22neurips} & 31.5 & 40.5 & 24.8 
        & 29.9 & 39.0 & 23.1 
        & 20.3 & 28.0 & 14.6 \\

    SigLIP 2~\cite{tschannen2025siglip} & ViT-B-16 @ 256
        &WebLI~\cite{chen2022pali}&
        27.6 & 36.9 & 20.7 &  
        24.2 & 32.5 & 18.0 &  
        9.9 & 14.7 & 6.3   \\

    SigLIP 2~\cite{tschannen2025siglip} & ViT-B-16 @ 384
        &WebLI~\cite{chen2022pali}&
       34.0 & 43.3 & 27.0 & 
        32.9 & 42.6 & 25.6 & 
        25.9 & 33.9 & 20.1 \\
    
     SigLIP 2~\cite{tschannen2025siglip} & ViT-B-16 @ 512
        &WebLI~\cite{chen2022pali}& 33.7 & 43.4 & 26.5 
        & 34.2 & 44.5 & 26.6 
        & 30.1 & 39.5 & 23.1 \\
    
    OpenCLIP~\cite{cherti23cvpr} & ConvNeXt-Base
        &LAION-2B~\cite{schuhmann22neurips}& 32.6 & 41.1 & 26.3 
        & 35.6 & 45.1 & 28.5 
        & 39.0 & 50.7 & 30.3 \\

    \midrule
    
    \rowcolor{gray!5} CLIP~\cite{radford21icml}     & ViT-L-14 @ 224 
        &WIT~\cite{radford21icml}& 12.5 & 19.3 &  7.4 
        & 13.4 & 19.8 &  8.6 
        & 13.1 & 20.8 &  7.4 \\
    \rowcolor{gray!5} OpenCLIP~\cite{cherti23cvpr} & ViT-L-14 @ 224
        &LAION-2B~\cite{schuhmann22neurips}& 10.8 & 15.9 &  7.0 
        & 11.2 & 16.7 &  7.0 
        & 10.7 & 16.3 &  6.5 \\
    \rowcolor{gray!5} SigLIP 2~\cite{tschannen2025siglip} & ViT-L-16 @ 256
        &WebLI~\cite{chen2022pali}
        & 34.9 & 42.4 & 29.3 & 
        31.2 & 37.8 & 26.3 & 
        18.7 & 24.8 & 14.2 \\
    \rowcolor{gray!5} SigLIP 2~\cite{tschannen2025siglip} & ViT-L-16 @ 384
        &WebLI~\cite{chen2022pali}
        &38.0 & 45.7 & 32.3 & 
        36.6 & 44.7 & 30.6 & 
       27.4 & 34.4 & 22.2  \\
    \rowcolor{gray!5} SigLIP 2~\cite{tschannen2025siglip} & ViT-L-16 @ 512
        &WebLI~\cite{chen2022pali}& 40.5 & 49.1 & 34.1 
        & 40.8 & 49.9 & 34.0 
        & 36.4 & 44.9 & 30.1 \\
    \rowcolor{gray!5} OpenCLIP~\cite{cherti23cvpr} & ConvNeXt-Large
        &LAION-2B~\cite{schuhmann22neurips}& 32.8 & 40.6 & 27.0 
        & 35.6 & 44.2 & 29.3 
        & 41.8 & 53.3 & 33.3 \\
    \bottomrule
  \end{tabular}
    \caption{
        Zero-shot 
        panoptic quality (PQ)
        of dense CLIP features 
        and our segmentation oracle
        on ADE20K val.
        We stratify classes on 
        ADE20K$\cap$COCO (seen) 
        and ADE20K\textbackslash COCO (unseen) according to~\cite{yu23neurips}.
    }
  \label{tab:seg_oracle}
\end{table*}
We first examine CLIP’s 
upper bound
as a zero-shot classifier
with oracle mask generator.
Our segmentation oracle produces
a set of perfect binary masks,
one for each panoptic segment. 
We use pre-trained 
CLIP encoders.
We extract dense image features 
in the shared vision-language 
embedding space
according to 
the MaskCLIP~\cite{zhou22eccv} strategy
for ViT backbones~\cite{dosovitskiy2020image}, 
while simply removing
the global average pool
for ConvNeXt~\cite{liu2022convnet}.
Then, we gather per-mask representations
by mask-pooling~\cite{xu2023open, yu23neurips} 
over the extracted CLIP features.
We embedd class names 
into the common vector space 
with the CLIP text encoder. 
Finally, we recover panoptic segmentation 
by assigning each mask with the class 
whose text embedding yields 
the highest cosine similarity 
with the corresponding 
visual representation. 
\Cref{fig:mask_pooling_oracle} illustrates the described procedure.

Table~\ref{tab:seg_oracle} 
presents results for
base (top) and large variants (bottom) of
ConvNeXt and ViT models. 
We evaluate three 
pre-trained ViT models: 
the original CLIP~\cite{radford21icml}, 
OpenCLIP~\cite{cherti23cvpr} 
and SigLIP2~\cite{tschannen2025siglip}. 
Each of these models
is tested
on three resolutions: 
longer side 512,
longer side 640, and 
shorter side 800 with 
longer side capped at 1333~\cite{yu23neurips}.
All experiments interpolate the ViT positional
embeddings to match the input resolution.
The experiments reveal 
several interesting insights,
which we discuss next.

First, ConvNeXt-Large achieves 
the highest performance 
among all evaluated models with 41.8 PQ. 
While impressive for 
a zero-shot setup, 
it falls nearly 
8 points short of 
our in-domain Mask2Former model 
with the same backbone (\cref{fig:teaser}). 
The gap is particularly striking
given that the vision-language models 
are evaluated with oracle segmentation boundaries. 
These findings indicate that, despite recent advances,
current VLMs still lack the dense perception
required for accurate panoptic segmentation.

Second, when comparing peak performance, we find that ConvNeXt still outperforms even the most recent ViT-based model, SigLIP2. 
At lower input resolutions, 
the base variants perform comparably, 
and SigLIP2-Large surpasses ConvNeXt-Large. 
However, ConvNeXt benefits 
markedly from the third 
input configuration
which employs larger 
and variable resolutions, 
while it 
degrades ViT models. 
Despite using training techniques that target dense prediction, SigLIP2 cannot match convolutional backbones. It does, however, achieve a substantial gain over earlier ViTs. This result confirms the value of its enhanced training strategy. 
The persistent gap suggests that ViTs 
still can not outperform 
the convolutional models 
at large resolutions.

Third, the performance is 
consistently lower 
on classes from ADE20K\textbackslash COCO ("unseen") 
than on those from ADE20K$\cap$COCO ("seen").
This gap is unexpected
as our model has not received
any training besides 
the CLIP pre-training.
The result suggests that 
the unseen subset 
in open-voc segmentation
experiments~\cite{yu23neurips}
is intrinsically harder.
A plausible explanation is class-frequency bias:
classes annotated in both datasets are likely more common in natural images,
and VLMs may favor such frequent concepts.

Finally, 
we analyze the impact 
of capacity 
onto panoptic performance. 
ConvNeXt benefits from larger models 
at higher input resolutions, 
while showing limited gains 
at lower scales. 
In contrast, SigLIP2 demonstrates consistent
and significant improvements 
across all resolutions 
as capacity increases. 
Moreover, MaskCLIP appears to extract unreliable
dense features from ViT-L/14,
which appears 
consistent with prior per-patch segmentation results~\cite{lan2024proxyclip}.

\researchfinding[segmentation oracle]{
\emph{\textbf{Finding 1:}} 
CLIP models struggle with 
region-level classification 
and fall short of in-domain baselines 
even when provided with perfect segmentation.
}

\subsection{Mask Classification Oracle}
\begin{table}[b!]
    \small
    \centering
    
    \begin{tabular}{lccc}
    \toprule
    Model & PQ\textsubscript{all} & PQ\textsubscript{seen} & PQ\textsubscript{unseen} \\
    \midrule
    FC-CLIP & 26.8 & 39.5 & 17.3 \\
    \blarrow \ + oracle classification & 39.8 & 47.5 & 34.2 \\
    \midrule
    MAFT+ & 26.9 & 37.0 & 19.5 \\
    \blarrow \ + oracle classification & 39.2 & 45.8 & 34.3 \\
    \bottomrule
    \end{tabular}
    \caption{Evaluating the impact of perfect mask classification on open-vocabulary panoptic segmentation on COCO$\rightarrow$ADE20K.}
    \label{tab:cls_oracle}
\end{table}

We now evaluate 
a classification oracle 
to assess the panoptic upper bound 
with the current mask proposals.
Our oracle adjusts 
the class posteriors
of the final panoptic map
for all masks
that overlap a ground truth 
segment with $\text{IoU} > 0.5$.

Table~\ref{tab:cls_oracle} presents 
the impact of oracle classification 
on open-vocabulary panoptic performance. 
Both FC-CLIP and MAFT+ 
show substantial gains 
of 13 PQ points overall
when provided with 
perfect classification. 
Specifically, 
PQ\textsubscript{seen} 
increases by 8.0 and 8.8 PQ points,
while we observe 
a twofold improvement in 
PQ\textsubscript{unseen}, 
16.9 and 14.8 PQ points 
for FC-CLIP and MAFT+ respectively.
These improvements 
confirm that 
recognition represents
a major component of 
the performance bottleneck.
This is particularly
the case
for unseen classes,
which may indicate 
overfitting on 
the training dataset.
However, even with 
perfect classification, 
the overall performance 
still falls short of 
typical in-domain baselines. 
Note that this oracle
can not correct
mistakes caused 
by the insufficient 
overlap between predicted 
and ground truth segments.
This suggests that 
the remaining limitations
lie in the quality 
of the mask proposals.

\researchfinding[segmentation oracle]{
\emph{\textbf{Finding 2:}} 
Oracle classification 
improves open-voc performance 
but still lags behind 
in-domain baselines, 
indicating significant shortcomings 
in mask proposals.
}
\subsection{Mask Selection Oracle}
To better understand 
the root causes of 
the observed limitations,
we dive deeper into the 
mask proposal generation.
Our classification oracle 
operates only 
on the set of masks included 
in the final 
panoptic prediction. 
However, before the  
prediction is assembled, 
the mask decoder 
typically produces 
a much larger set of candidates
consisting of up to $N = 250$ masks.
This raises a key question: 
is the mask selection process 
that reduces 
this candidate set optimal?

To explore this, 
we conduct an experiment 
using an oracle 
mask selection. 
Specifically, 
we apply Hungarian matching 
between the ground truth masks
and the full set of candidate masks 
to identify those that 
best explain the ground truth. 
The matching cost is 
computed using a combination 
of binary cross-entropy 
and Dice loss. 
After matching, 
we discard the unmatched 
candidate masks and proceed 
with the standard panoptic inference~\cite{cheng22cvpr}.
Note that this oracle 
is relatively non-intrusive, 
as it does not directly alter 
classification or segmentation, 
but solely influences 
the selection of masks 
from those already generated.
Table~\ref{tab:oracle_matching}
presents the results.

\begin{table}[h]
    \centering
    \small
    \begin{tabular}{@{}l@{\quad}cc@{\enspace}c@{}}
    \toprule
    Model & PQ\textsubscript{all} & PQ\textsubscript{seen} & PQ\textsubscript{unseen} \\
    \midrule
    FC-CLIP & 26.8 & 39.5 & 17.3 \\
    \blarrow \  + oracle mask selection & 21.9 & 33.6 & 13.2 \\
    \quad \blarrow \  + dropping "no-object" logit  & 36.7 & 48.9 & 27.7 \\
    \midrule
    MAFT+ & 26.9 & 37.0 & 17.4 \\
    \blarrow \  + oracle mask selection  & 25.4 & 35.7 & 17.6 \\
    \quad \blarrow\ + dropping "no-object" logit & 33.7 & 42.6 & 27.1 \\
    \bottomrule
    \end{tabular}
    \caption{Evaluating the impact of oracle mask-selection on open-vocabulary panoptic segmentation on COCO$\rightarrow$ADE20K. 
    }
    \label{tab:oracle_matching}
\end{table}
Surprisingly, 
oracle mask selection (\cf row 2)
causes a significant
performance drop 
of 4.9 PQ points 
and 1.5 PQ points 
for FC-CLIP 
and MAFT+.
This leads us to visually
inspect and compare
regular predictions
to those with 
oracle mask selection.
We notice that 
in some cases
oracle selection
causes disappearance
of correct masks
due to the filtering 
of masks with
no-object posterior. 
Specifically, 
if the classifier assigns 
the highest probability 
to the special "no-object" class, 
the corresponding mask is discarded 
and excluded from 
the final panoptic segmentation. 
We therefore modify 
the inference by removing 
the "no-object" logit, 
ensuring that all 
oracle-selected masks 
are included 
in panoptic prediction.
The third row 
of each section
in Table~\ref{tab:oracle_matching}
presents experiments
with the modified selection oracle.
We observe a 
significant performance boost
of 9.9 and 6.8 PQ points
for FC-CLIP and MAFT+.
These findings are 
particularly compelling, 
as they suggest that 
open-vocabulary models 
could achieve significant gains 
through more effective 
mask selection alone.
The improvement is especially 
notable on unseen classes, 
implying that the models
internally often succeed 
in localizing these objects 
as well as assigning
a correct semantic category.
This indicates that
many accurate masks
get discarded
due to poorly calibrated 
no-object embedding
or being eclipsed 
by other masks.
We present a more detailed analysis
of this phenomena
in~\ref{sec:in-domain-comparison}.
\researchfinding[segmentation oracle]{
\emph{\textbf{Finding 3:}} Many valid mask proposals are discarded due to the incorrect classification as \texttt{no-object}.
}

Table~\ref{tab:oracle_matching2} 
extends the analysis 
of the mask-selection oracle 
by pairing it with either 
oracle segmentation (row 2) 
or oracle classification (row 3). 
The segmentation oracle 
uses ground truth 
to fix the boundaries
of the matched masks,
while retaining the 
mask classifications 
of the corresponding 
open-voc segmentation model.
We observe improvements 
of 13.6 and 18.7 PQ points 
over the mask-selection oracle 
for FC-CLIP and MAFT+, respectively.
Notably, the gains are larger for the seen classes, 
suggesting that the models struggle more
with classifying unseen categories. 
Additional insights emerge 
when comparing these results 
with those in Table~\ref{tab:seg_oracle}, 
which also evaluates the segmentation oracle 
but in combination with the zero-shot VLMs.
We observe that both 
FC-CLIP and MAFT+ 
outperform the best VLMs 
in presence of segmentation oracle.
Interestingly, they also achieve 
higher performance 
on the unseen classes. 
This indicates that VLMs 
can benefit from ensembling 
with a trained mask classifier (as in FC-CLIP) 
or from fine-tuning of the visual backbone (as in MAFT+), 
even for the recognition of classes 
not present in the training set.

The classification oracle extends 
the optimal mask selection 
by assigning correct 
one-hot class probabilities 
based on the matching 
with ground truth segments. 
This oracle raises 
the performance upper bound 
well beyond the 
in-domain models.
Specifically, FC-CLIP reaches 66.4 PQ, 
while MAFT+ follows closely with 58.1 PQ. 
Interestingly, performance on seen 
and unseen classes is comparable
in this setting. 
This supports our 
earlier observation 
that the mask proposal generator 
tends to discard 
valid mask candidates 
for unseen classes, 
and that the classifier 
particularly struggles
with these instances.
The remaining performance gap 
suggests that many ground truth 
segments still 
lack appropriate
corresponding masks, 
even with oracle mask selection. 
Although the matching process 
identifies the optimal candidate mask 
for each ground truth segment, 
it does not guarantee sufficient overlap 
to ensure a correct match.
\begin{table}[h]
    \centering
    \small
    \begin{tabular}{ccccc}
    \toprule
    + Oracle Seg. & + Oracle Cls. & PQ\textsubscript{all} & PQ\textsubscript{seen} & PQ\textsubscript{unseen} \\
    \midrule
    \multicolumn{5}{c}{{FC-CLIP} w/ oracle mask selection} \\
    \hline
    \n & \n & 36.7 & 48.9 & 27.7 \\
    \y & \n & 50.3 & 66.0 & 38.7 \\
    \n & \y & 66.4 & 67.1 & 65.9 \\
    \toprule
    \multicolumn{5}{c}{{MAFT+} w/ oracle mask selection} \\
    \hline
    \n & \n & 33.7 & 42.6 & 27.1 \\
    \y & \n & 52.4 & 63.0 & 44.4 \\
    \n & \y & 58.1 & 59.8 & 56.8 \\
    \bottomrule
    \end{tabular}
    \caption{Evaluating the impact of oracle segmentation and classification on open-vocabulary performance on COCO$\rightarrow$ADE20K.}
    \label{tab:oracle_matching2}
\end{table}
\researchfinding[segmentation oracle]{
\emph{\textbf{Finding 4:}} 
Oracle mask proposal selection and 
oracle mask classification 
boost performance well beyond in-domain baselines, highlighting these two components as the key limitations in current open-vocabulary models.
}
\subsection{Comparisons with In-domain Models}
\label{sec:in-domain-comparison}
\begin{figure}[h!]
    \centering
    \includegraphics[width=\linewidth]{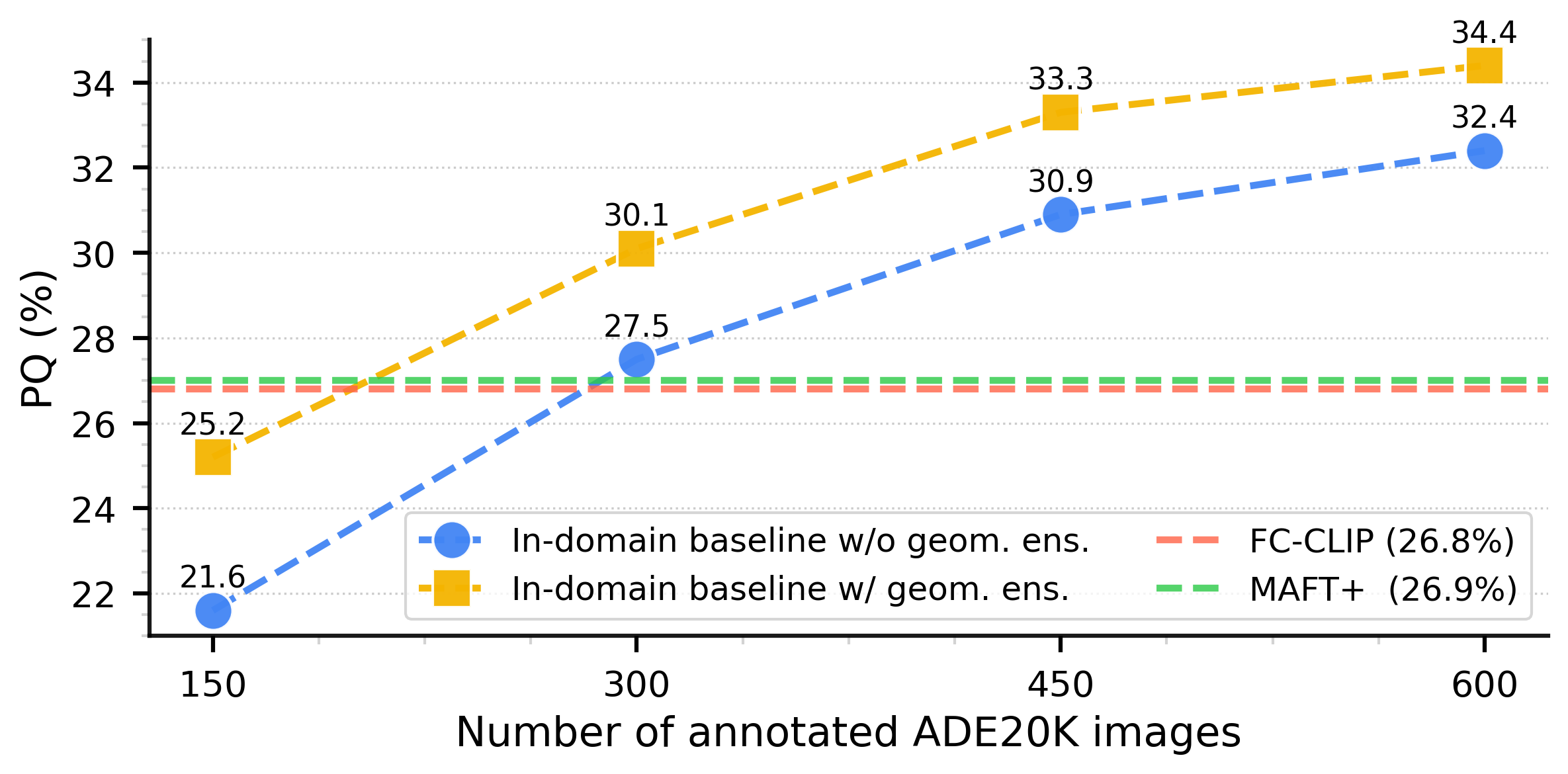}
    \caption{
        Performance comparison of open-vocabulary models MAFT+ (green) and FC-CLIP (red) with in-domain models trained on limited supervised data, evaluated either with (yellow) or without (blue) geometric ensembling with CLIP predictions.
    }
    \label{fig:supervised_comparison}
\end{figure}

We complete our analysis 
with a more detailed performance comparison 
between open-vocabulary 
and in-domain models. Figure~\ref{fig:supervised_comparison} 
aims to evaluate 
the generalization capabilities 
of open-vocabulary models 
and assess 
the impact of domain shift 
on their performance. 
Specifically, 
it presents the in-domain performance 
of FC-CLIP when trained 
on small subsets of 
ADE20K train 
and evaluated on 
ADE20K val, 
both with (yellow) 
and without (blue) 
geometric ensembling. 
For comparison, 
we also include 
the performance of FC-CLIP (red) 
and MAFT+ (green)
trained on the full 
COCO training set. 
The results show that the 
in-domain model matches 
the performance of open-vocabulary models 
using as few as 300 labeled images. 
With geometric ensembling, 
it even surpasses them by 3 PQ points. 
These findings raise questions 
about the practical value 
of open-vocabulary evaluation 
when comparable performance 
can be achieved within 
a target domain at 
such a low annotation cost. 
This suggests that current 
open-vocabulary models still struggle
with domain shift.
However, 
the domain shift
between COCO and ADE20K 
in terms of image appearance
should not be
particularly pronounced, 
as both datasets are web-scraped 
from Flickr and similar sources.
Hence, we argue that there 
must be some sort of 
labeling policy shift,
which represents
an insurmountable obstacle
for current
open-vocabulary models.

To examine this more closely, 
we identify classes 
where the open-vocabulary model
achieves a recall rate below 10\%,
and exhibits the largest drop 
in true positives 
compared to the in-domain model.
Table~\ref{tab:hard_cls}
lists top five such classes,
though many others 
are similarly affected.
The table also 
shows the number
of true positive
and false negative 
segments.
We further distinguish 
between false negatives 
caused by the absence 
of a predicted mask 
with sufficient segmentation overlap (\fnseg), 
and those where a correct mask 
was present but the predicted 
class was incorrect (\fncls).
We observe that the ratio 
of true positives to 
the total number of 
ground truth segments (FN + TP) is negligible, 
indicating a near-complete failure 
in recognizing these classes. 
The stratification of false negatives 
further reveals that most of the errors 
arise from inadequate segmentation
rather than misclassification.
\begin{table}[t]
    \centering
    \small
    \begin{tabular}{lr@{\;}r@{\;}rr@{\;}r@{\;}r}
    \toprule
    & \multicolumn{3}{c}{FC-CLIP~\cite{yu23neurips}} & \multicolumn{3}{c}{MAFT+~\cite{jiao24eccv}}\\
    \cmidrule(lr){2-4}  \cmidrule(lr){5-7}
    class name & \fnseg{} & \fncls{} & \tp{} & \fnseg{} & \fncls{} & \tp{} \\
    \toprule
    \texttt{light} & 1213 & 27 & 0 & 1212 & 16 & 12 \\
    \texttt{painting} & 708 & 38 & 40 & 756 & 20 & 10 \\
    \texttt{cushion} & 458 & 16 & 8 & 472 & 7 & 3 \\
    \texttt{sign} & 700 & 21 & 9 & 706 & 16 & 8 \\
    \texttt{pillow} & 241 & 0 & 2 & 240 & 1 & 2 \\
    \bottomrule
    \end{tabular}
    \caption{
        Classes that suffer the most in transition between the in-domain and open-vocabulary setups.
        We show true positives (\tp{}),
        and false negatives caused either by 
        misssegmentation (\fnseg{})
        or missclassification (\fncls{}).
    }
    \label{tab:hard_cls}
\end{table}

Manual inspection 
of COCO and ADE20K annotations 
reveals labeling conflicts 
for all classes from~\Cref{tab:hard_cls}. 
These conflicts are 
illustrated in Figure~\ref{fig:labconflicts}. 
The first three columns 
show ADE20K images overlaid 
with panoptic maps 
from the ground truth, 
or predictions of
FC-CLIP and MAFT+, respectively. 
The final column presents COCO images 
with overlaid panoptic maps 
that highlight the specific labeling conflict illustrated in each row. 
For clarity, some classes 
are omitted from the visualization.
The first row illustrates 
the discrepancy in labeling 
paintings on walls. 
In ADE20K, paintings are labeled
as a distinct thing class, 
\texttt{painting, picture}, 
whereas in COCO, 
they are most often left unlabeled 
or occasionally treated 
as part of the surrounding 
wall segments. 
As a result, both FC-CLIP and MAFT+ 
fail to recognize paintings in ADE20K. 
This failure stems 
from the limitations 
of the mask proposal generator, 
which is trained with a supervision 
that directly contradicts 
the evaluation objectives. 
Specifically, since paintings 
are not labeled in COCO, 
the model is never encouraged 
to generate masks that 
cover painting regions. 
If such a mask is 
proposed by chance, 
the model is trained 
to classify it as \texttt{no-object}. 
Consequently, most of these masks are 
discarded before the VLM even has 
an opportunity to classify 
them as paintings.
The second row 
considers the class \texttt{pillow},
which is present 
in the taxonomies of both datasets.
However, COCO excludes sleeping 
pillows on beds
and labels them as part of the bed.
On the other hand,
ADE20K uses \texttt{pillow} 
specifically for sleeping pillows 
and labels other 
types as \texttt{cushion}. 
As a result, open-vocabulary models
trained on COCO fail 
to recognize bed pillows 
in ADE20K, as shown in 
the first two rows.
This example illustrates 
the ambiguity of class names
and the necessity
for more accurate 
class descriptions
in open-vocabulary 
segmentation.
This also highlights 
another limitation 
of current 
open-voc models: 
they struggle to recognize 
classes that are subparts 
of training categories, 
as mask proposal generators 
are trained to produce 
a single mask for 
the whole object.
The third row presents 
an example for the class \texttt{signboard, sign}, which in ADE20K includes traffic signs. 
In contrast, 
COCO provides a dedicated label 
only for stop signs, 
while the other traffic signs 
are often left unlabeled 
or annotated as part 
of surrounding objects, 
such as buildings. 
This inconsistency leads 
to similar recognition issues 
as before.
\researchfinding[segmentation oracle]{
\emph{\textbf{Finding 5:}} 
Annotation conflicts between COCO and ADE20K reveal a misalignment between training supervision and evaluation objectives, causing open-vocabulary models to discard valid mask proposals.
}
\newcommand{\myw}{0.49\columnwidth}
\begin{figure*}[t]
    \centering
    \begin{tabular}{@{}c@{\,}c@{\,}c@{\,}c@{}}
         \footnotesize ADE20K annotations & \footnotesize FC-CLIP~\cite{yu23neurips} prediction & \footnotesize MAFT+~\cite{jiao24eccv} prediction & \footnotesize COCO annotations \\
         \includegraphics[width=\myw]{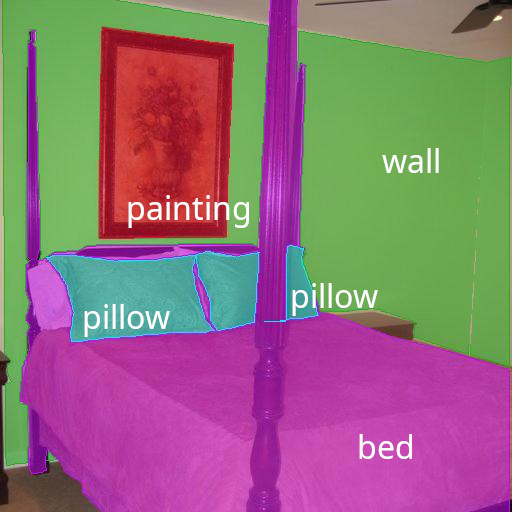} & 
         \includegraphics[width=\myw]{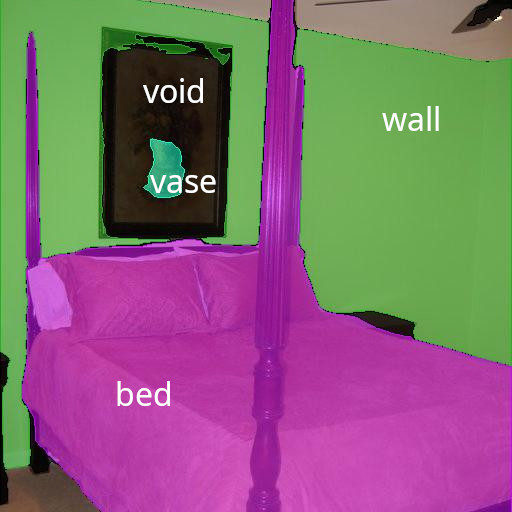} & 
         \includegraphics[width=\myw]{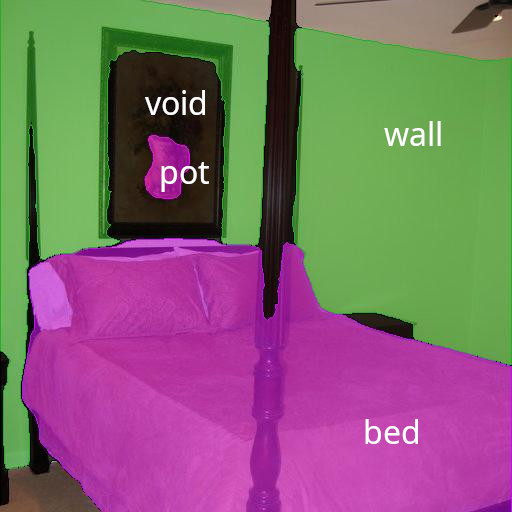} & 
         \includegraphics[width=\myw]{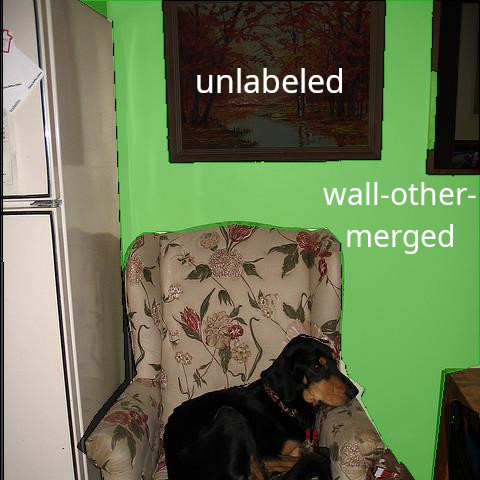} \\[-0.2em]
         \includegraphics[width=\myw]{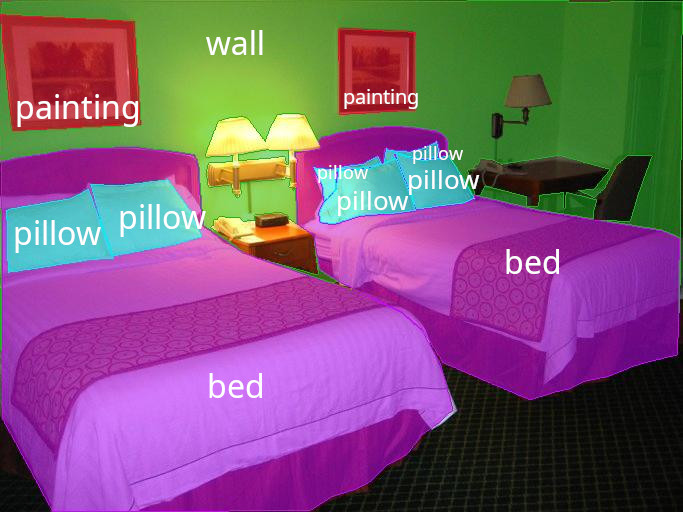} & 
         \includegraphics[width=\myw]{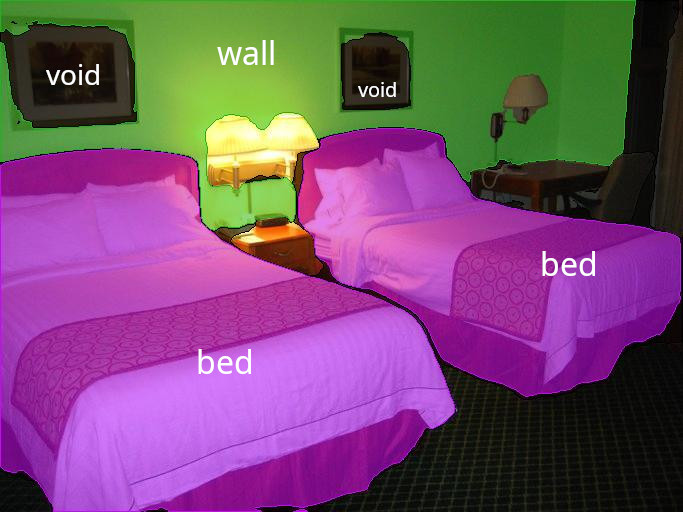} & 
         \includegraphics[width=\myw]{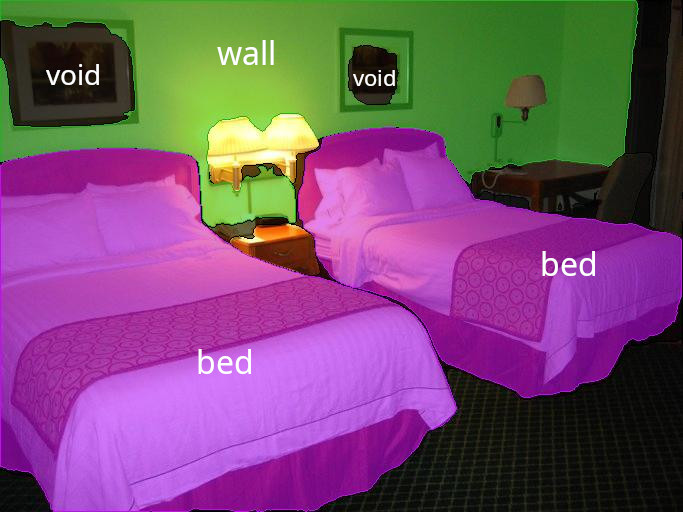} & 
         \includegraphics[width=\myw]{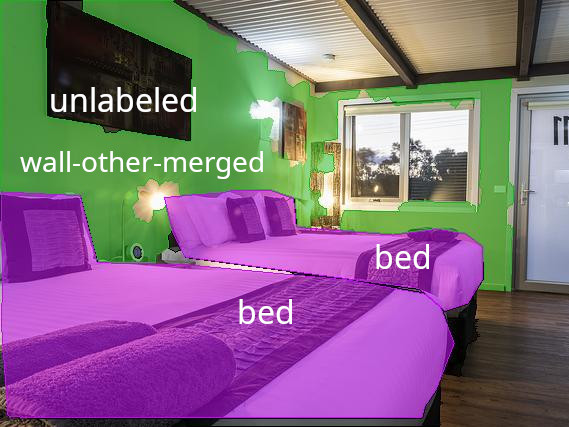} \\[-0.2em]
         \includegraphics[width=\myw]{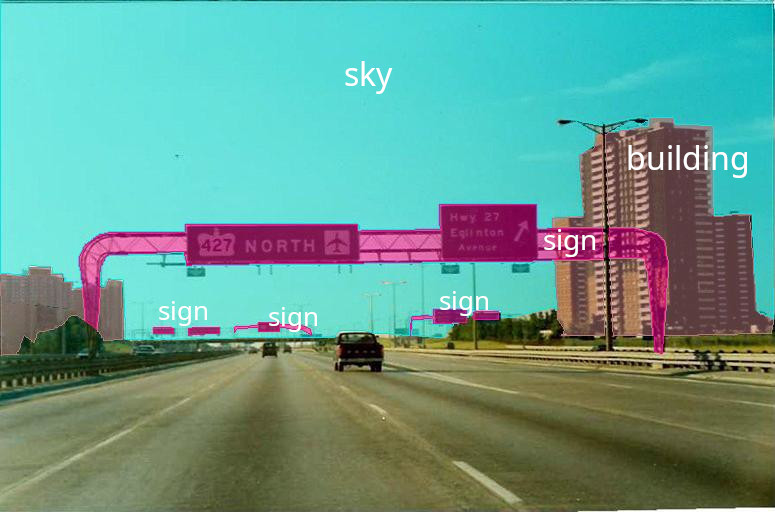} & 
         \includegraphics[width=\myw]{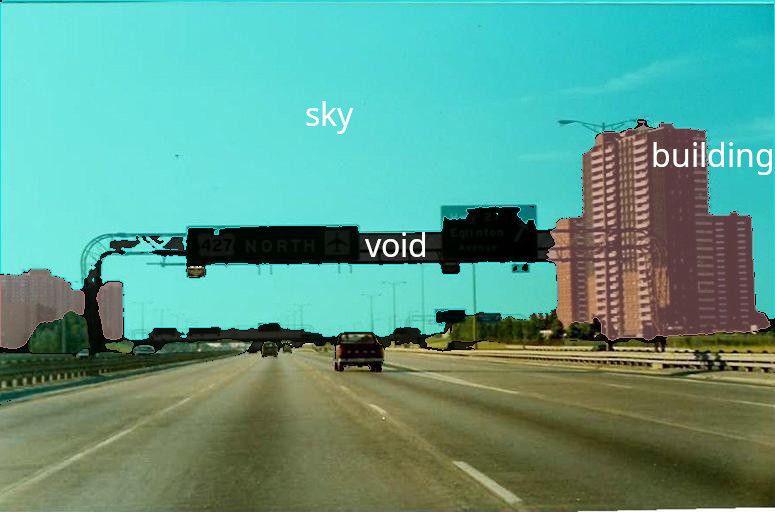} & 
         \includegraphics[width=\myw]{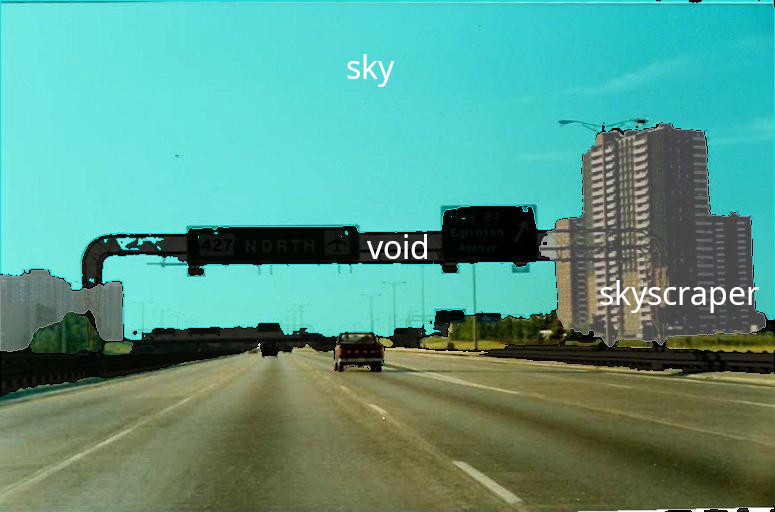} & 
         \includegraphics[width=\myw]{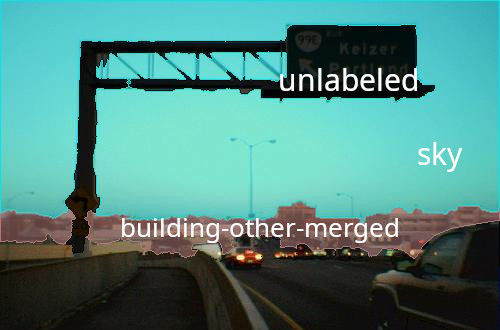} \\
    \end{tabular}
    \caption{
        Illustration of labeling policy conflicts between COCO and ADE20K that hinder open-vocabulary model performance. The rows illustrate labeling conflicts for classes \texttt{painting}, \texttt{pillow}, and traffic \texttt{sign}, respectively.
    }
    \label{fig:labconflicts}
\end{figure*}

\section{Conclusion}
\label{sec:conclusion}
We have presented a comprehensive analysis
of mask-transformer methods for 
open-vocabulary segmentation. 
Our oracle experiments identify the key
bottlenecks in both panoptic subtasks:
mask classification and segmentation.
First, the current vision–language
models struggle with region-level classification:
even with perfect masks, 
they lag behind in-domain 
performance.
This gap underscores the
need for better VLM pre-training
pipelines to enhance their
dense representations.
Second, our analysis shows that
even with oracle classification,
prominent open-vocabulary approaches
still lag behind in-domain models,
revealing shortcomings
in mask proposal generation.
Third, we demonstrate
that mask proposal generators
internally produce valid proposals,
yet discard them at inference
due to biases inherited 
from the training data.
These biases prevent them from
reconciling labeling-policy
discrepancies between 
training and evaluation datasets.

The identified bottlenecks
suggest several promising
research directions.
First, 
future work 
should eliminate 
the taxonomy conflicts 
by unifying or precisely mapping label sets 
in benchmark design to ensure fair
and reliable evaluation 
of open-vocabulary performance.
Second, current proposal
generators lack vocabulary awareness:
they generate the same mask candidates
regardless of the test-time taxonomy.
Future work should develop 
vocabulary-aware proposal generators
that dynamically adapt mask boundaries
to the evaluation vocabulary.
Finally, we argue that richer
annotation guidance is essential
and can be achieved in two ways:
i) in a few-shot setting,
by providing exemplar segmentations
that illustrate the desired outputs, and
ii) by supplying detailed textual guidelines
that define annotation rules 
for each semantic category. 
The latter approach appears more feasible,
as articulating annotation guidelines
in natural language tends
to be easier than collecting
examples that cover every edge case.
However, language-based approach
could be limited because the
current CLIP text encoders
operate largely as a bag-of-words. 
Replacing these encoders
with large language 
models could
capture finer
class distinctions,
but only
if the vision encoders can
match such subtle differences.
We believe 
these directions
offer a roadmap
toward more \textit{open}
open-vocab segmentation.

\section*{Acknowledgments}
This work has been supported
by Croatian Recovery
and Resilience Fund -
NextGenerationEU (grant C1.4 R5-I2.01.0001),
Advanced computing service provided
by the University of Zagreb
University Computing Centre - SRCE,
Slovenian research agency research program \mbox{P2-0214},
and European Union's Horizon Europe
research and innovation programme under the Marie Skłodowska-Curie Postdoctoral Fellowship Programme, SMASH co-funded under the grant agreement No.~101081355.
The SMASH project is co-funded by
the Republic of Slovenia and
the European Union from the
European Regional Development Fund.
In memory of dear mentor, colleague and friend, Siniša Šegvić.

{
    \small
    \bibliographystyle{ieeenat_fullname}
    \bibliography{main}

\begin{thebibliography}{65}
\providecommand{\natexlab}[1]{#1}
\providecommand{\url}[1]{\texttt{#1}}
\expandafter\ifx\csname urlstyle\endcsname\relax
  \providecommand{\doi}[1]{doi: #1}\else
  \providecommand{\doi}{doi: \begingroup \urlstyle{rm}\Url}\fi

\bibitem[Cha et~al.(2023)Cha, Mun, and Roh]{cha2023learning}
Junbum Cha, Jonghwan Mun, and Byungseok Roh.
\newblock Learning to generate text-grounded mask for open-world semantic segmentation from only image-text pairs.
\newblock In \emph{Proceedings of the IEEE/CVF Conference on Computer Vision and Pattern Recognition}, pages 11165--11174, 2023.

\bibitem[Chen et~al.(2017)Chen, Papandreou, Kokkinos, Murphy, and Yuille]{chen17pami}
Liang-Chieh Chen, George Papandreou, Iasonas Kokkinos, Kevin Murphy, and Alan~L Yuille.
\newblock Deeplab: Semantic image segmentation with deep convolutional nets, atrous convolution, and fully connected crfs.
\newblock \emph{IEEE transactions on pattern analysis and machine intelligence}, 40\penalty0 (4):\penalty0 834--848, 2017.

\bibitem[Chen et~al.(2018)Chen, Zhu, Papandreou, Schroff, and Adam]{chen2018encoder}
Liang-Chieh Chen, Yukun Zhu, George Papandreou, Florian Schroff, and Hartwig Adam.
\newblock Encoder-decoder with atrous separable convolution for semantic image segmentation.
\newblock In \emph{Proceedings of the European conference on computer vision (ECCV)}, pages 801--818, 2018.

\bibitem[Chen et~al.(2022)Chen, Wang, Changpinyo, Piergiovanni, Padlewski, Salz, Goodman, Grycner, Mustafa, Beyer, et~al.]{chen2022pali}
Xi Chen, Xiao Wang, Soravit Changpinyo, AJ Piergiovanni, Piotr Padlewski, Daniel Salz, Sebastian Goodman, Adam Grycner, Basil Mustafa, Lucas Beyer, et~al.
\newblock Pali: A jointly-scaled multilingual language-image model.
\newblock \emph{arXiv preprint arXiv:2209.06794}, 2022.

\bibitem[Cheng et~al.(2021)Cheng, Schwing, and Kirillov]{cheng21neurips}
Bowen Cheng, Alex Schwing, and Alexander Kirillov.
\newblock Per-pixel classification is not all you need for semantic segmentation.
\newblock \emph{Advances in neural information processing systems}, 34:\penalty0 17864--17875, 2021.

\bibitem[Cheng et~al.(2022)Cheng, Misra, Schwing, Kirillov, and Girdhar]{cheng22cvpr}
Bowen Cheng, Ishan Misra, Alexander~G. Schwing, Alexander Kirillov, and Rohit Girdhar.
\newblock Masked-attention mask transformer for universal image segmentation.
\newblock In \emph{Proceedings of the IEEE/CVF Conference on Computer Vision and Pattern Recognition (CVPR)}, pages 1290--1299, 2022.

\bibitem[Cherti et~al.(2023)Cherti, Beaumont, Wightman, Wortsman, Ilharco, Gordon, Schuhmann, Schmidt, and Jitsev]{cherti23cvpr}
Mehdi Cherti, Romain Beaumont, Ross Wightman, Mitchell Wortsman, Gabriel Ilharco, Cade Gordon, Christoph Schuhmann, Ludwig Schmidt, and Jenia Jitsev.
\newblock Reproducible scaling laws for contrastive language-image learning.
\newblock In \emph{Proceedings of the IEEE/CVF Conference on Computer Vision and Pattern Recognition}, pages 2818--2829, 2023.

\bibitem[Cho et~al.(2024)Cho, Shin, Hong, Arnab, Seo, and Kim]{cho2024cat}
Seokju Cho, Heeseong Shin, Sunghwan Hong, Anurag Arnab, Paul~Hongsuck Seo, and Seungryong Kim.
\newblock Cat-seg: Cost aggregation for open-vocabulary semantic segmentation.
\newblock In \emph{Proceedings of the IEEE/CVF Conference on Computer Vision and Pattern Recognition}, pages 4113--4123, 2024.

\bibitem[Cordts et~al.(2016)Cordts, Omran, Ramos, Rehfeld, Enzweiler, Benenson, Franke, Roth, and Schiele]{cordts16cvpr}
Marius Cordts, Mohamed Omran, Sebastian Ramos, Timo Rehfeld, Markus Enzweiler, Rodrigo Benenson, Uwe Franke, Stefan Roth, and Bernt Schiele.
\newblock The cityscapes dataset for semantic urban scene understanding.
\newblock In \emph{Proceedings of the IEEE conference on computer vision and pattern recognition}, pages 3213--3223, 2016.

\bibitem[Demir et~al.(2018)Demir, Koperski, Lindenbaum, Pang, Huang, Basu, Hughes, Tuia, and Raskar]{demir18cvprw}
Ilke Demir, Krzysztof Koperski, David Lindenbaum, Guan Pang, Jing Huang, Saikat Basu, Forest Hughes, Devis Tuia, and Ramesh Raskar.
\newblock Deepglobe 2018: A challenge to parse the earth through satellite images.
\newblock In \emph{Proceedings of the IEEE conference on computer vision and pattern recognition workshops}, pages 172--181, 2018.

\bibitem[Ding et~al.(2022)Ding, Xue, Xia, and Dai]{ding2022decoupling}
Jian Ding, Nan Xue, Gui-Song Xia, and Dengxin Dai.
\newblock Decoupling zero-shot semantic segmentation.
\newblock In \emph{Proceedings of the IEEE/CVF conference on computer vision and pattern recognition}, pages 11583--11592, 2022.

\bibitem[Dosovitskiy et~al.(2020)Dosovitskiy, Beyer, Kolesnikov, Weissenborn, Zhai, Unterthiner, Dehghani, Minderer, Heigold, Gelly, et~al.]{dosovitskiy2020image}
Alexey Dosovitskiy, Lucas Beyer, Alexander Kolesnikov, Dirk Weissenborn, Xiaohua Zhai, Thomas Unterthiner, Mostafa Dehghani, Matthias Minderer, Georg Heigold, Sylvain Gelly, et~al.
\newblock An image is worth 16x16 words: Transformers for image recognition at scale.
\newblock \emph{arXiv preprint arXiv:2010.11929}, 2020.

\bibitem[Everingham et~al.(2010)Everingham, Van~Gool, Williams, Winn, and Zisserman]{everingham2010pascal}
Mark Everingham, Luc Van~Gool, Christopher~KI Williams, John Winn, and Andrew Zisserman.
\newblock The pascal visual object classes (voc) challenge.
\newblock \emph{International journal of computer vision}, 88:\penalty0 303--338, 2010.

\bibitem[Fang et~al.(2024)Fang, Jose, Jain, Schmidt, Toshev, and Shankar]{fang2024data}
Alex Fang, Albin~Madappally Jose, Amit Jain, Ludwig Schmidt, Alexander~T Toshev, and Vaishaal Shankar.
\newblock Data filtering networks.
\newblock In \emph{The Twelfth International Conference on Learning Representations}, 2024.

\bibitem[Ghiasi et~al.(2022)Ghiasi, Gu, Cui, and Lin]{ghiasi22eccv}
Golnaz Ghiasi, Xiuye Gu, Yin Cui, and Tsung-Yi Lin.
\newblock Scaling open-vocabulary image segmentation with image-level labels.
\newblock In \emph{European conference on computer vision}, pages 540--557. Springer, 2022.

\bibitem[Girshick(2015)]{girshick15iccv}
Ross Girshick.
\newblock Fast r-cnn.
\newblock In \emph{Proceedings of the IEEE international conference on computer vision}, pages 1440--1448, 2015.

\bibitem[Girshick et~al.(2014)Girshick, Donahue, Darrell, and Malik]{girshick14cvpr}
Ross Girshick, Jeff Donahue, Trevor Darrell, and Jitendra Malik.
\newblock Rich feature hierarchies for accurate object detection and semantic segmentation.
\newblock In \emph{Proceedings of the IEEE conference on computer vision and pattern recognition}, pages 580--587, 2014.

\bibitem[Han et~al.(2023)Han, Zhong, Li, Han, and Ma]{han2023deop}
Cong Han, Yujie Zhong, Dengjie Li, Kai Han, and Lin Ma.
\newblock Open-vocabulary semantic segmentation with decoupled one-pass network.
\newblock In \emph{Proceedings of the IEEE/CVF International Conference on Computer Vision (ICCV)}, pages 1086--1096, 2023.

\bibitem[Hariharan et~al.(2014)Hariharan, Arbel{\'a}ez, Girshick, and Malik]{hariharan14eccv}
Bharath Hariharan, Pablo Arbel{\'a}ez, Ross Girshick, and Jitendra Malik.
\newblock Simultaneous detection and segmentation.
\newblock In \emph{Computer Vision--ECCV 2014: 13th European Conference, Zurich, Switzerland, September 6-12, 2014, Proceedings, Part VII 13}, pages 297--312. Springer, 2014.

\bibitem[He et~al.(2017)He, Gkioxari, Doll{\'a}r, and Girshick]{he17iccv}
Kaiming He, Georgia Gkioxari, Piotr Doll{\'a}r, and Ross Girshick.
\newblock Mask r-cnn.
\newblock In \emph{Proceedings of the IEEE international conference on computer vision}, pages 2961--2969, 2017.

\bibitem[Jia et~al.(2021)Jia, Yang, Xia, Chen, Parekh, Pham, Le, Sung, Li, and Duerig]{jia21icml}
Chao Jia, Yinfei Yang, Ye Xia, Yi-Ting Chen, Zarana Parekh, Hieu Pham, Quoc Le, Yun-Hsuan Sung, Zhen Li, and Tom Duerig.
\newblock Scaling up visual and vision-language representation learning with noisy text supervision.
\newblock In \emph{International conference on machine learning}, pages 4904--4916. PMLR, 2021.

\bibitem[Jiao et~al.(2024)Jiao, Zhu, Huang, Zhao, Wei, and Shi]{jiao24eccv}
Siyu Jiao, Hongguang Zhu, Jiannan Huang, Yao Zhao, Yunchao Wei, and Humphrey Shi.
\newblock Collaborative vision-text representation optimizing for open-vocabulary segmentation.
\newblock In \emph{European Conference on Computer Vision}, pages 399--416. Springer, 2024.

\bibitem[Karazija et~al.(2024)Karazija, Laina, Vedaldi, and Rupprecht]{karazija2024diffusion}
Laurynas Karazija, Iro Laina, Andrea Vedaldi, and Christian Rupprecht.
\newblock Diffusion models for open-vocabulary segmentation.
\newblock In \emph{European Conference on Computer Vision}, pages 299--317. Springer, 2024.

\bibitem[Kirillov et~al.(2019)Kirillov, He, Girshick, Rother, and Doll{\'a}r]{kirillov19cvpr}
Alexander Kirillov, Kaiming He, Ross Girshick, Carsten Rother, and Piotr Doll{\'a}r.
\newblock Panoptic segmentation.
\newblock In \emph{Proceedings of the IEEE/CVF conference on computer vision and pattern recognition}, pages 9404--9413, 2019.

\bibitem[kokitsi Maninis et~al.(2025)kokitsi Maninis, Chen, Ghosh, Karpur, Chen, Xia, Cao, Salz, Han, Dlabal, Gnanapragasam, Seyedhosseini, Zhou, and Araujo]{maninis2025tips}
Kevis kokitsi Maninis, Kaifeng Chen, Soham Ghosh, Arjun Karpur, Koert Chen, Ye Xia, Bingyi Cao, Daniel Salz, Guangxing Han, Jan Dlabal, Dan Gnanapragasam, Mojtaba Seyedhosseini, Howard Zhou, and Andre Araujo.
\newblock {TIPS}: Text-image pretraining with spatial awareness.
\newblock In \emph{The Thirteenth International Conference on Learning Representations}, 2025.

\bibitem[Kreso et~al.(2017)Kreso, Segvic, and Krapac]{kreso2017ladder}
Ivan Kreso, Sinisa Segvic, and Josip Krapac.
\newblock Ladder-style densenets for semantic segmentation of large natural images.
\newblock In \emph{Proceedings of the IEEE International Conference on Computer Vision Workshops}, pages 238--245, 2017.

\bibitem[Lan et~al.(2024{\natexlab{a}})Lan, Chen, Ke, Wang, Feng, and Zhang]{lan2024clearclip}
Mengcheng Lan, Chaofeng Chen, Yiping Ke, Xinjiang Wang, Litong Feng, and Wayne Zhang.
\newblock Clearclip: Decomposing clip representations for dense vision-language inference.
\newblock In \emph{European Conference on Computer Vision}, pages 143--160. Springer, 2024{\natexlab{a}}.

\bibitem[Lan et~al.(2024{\natexlab{b}})Lan, Chen, Ke, Wang, Feng, and Zhang]{lan2024proxyclip}
Mengcheng Lan, Chaofeng Chen, Yiping Ke, Xinjiang Wang, Litong Feng, and Wayne Zhang.
\newblock Proxyclip: Proxy attention improves clip for open-vocabulary segmentation.
\newblock In \emph{European Conference on Computer Vision}, pages 70--88. Springer, 2024{\natexlab{b}}.

\bibitem[Li et~al.(2023{\natexlab{a}})Li, Zhang, Xu, Liu, Zhang, Ni, and Shum]{li23cvpr}
Feng Li, Hao Zhang, Huaizhe Xu, Shilong Liu, Lei Zhang, Lionel~M Ni, and Heung-Yeung Shum.
\newblock Mask dino: Towards a unified transformer-based framework for object detection and segmentation.
\newblock In \emph{Proceedings of the IEEE/CVF conference on computer vision and pattern recognition}, pages 3041--3050, 2023{\natexlab{a}}.

\bibitem[Li et~al.(2023{\natexlab{b}})Li, Wang, and Xie]{li2023inverse}
Xianhang Li, Zeyu Wang, and Cihang Xie.
\newblock An inverse scaling law for clip training.
\newblock \emph{Advances in Neural Information Processing Systems}, 36:\penalty0 49068--49087, 2023{\natexlab{b}}.

\bibitem[Li et~al.(2023{\natexlab{c}})Li, Fan, Hu, Feichtenhofer, and He]{li2023scaling}
Yanghao Li, Haoqi Fan, Ronghang Hu, Christoph Feichtenhofer, and Kaiming He.
\newblock Scaling language-image pre-training via masking.
\newblock In \emph{Proceedings of the IEEE/CVF conference on computer vision and pattern recognition}, pages 23390--23400, 2023{\natexlab{c}}.

\bibitem[Li et~al.(2023{\natexlab{d}})Li, Wang, Duan, and Li]{li2023clipsurgery}
Yi Li, Hualiang Wang, Yiqun Duan, and Xiaomeng Li.
\newblock Clip surgery for better explainability with enhancement in open-vocabulary tasks.
\newblock \emph{arXiv e-prints}, pages arXiv--2304, 2023{\natexlab{d}}.

\bibitem[Li et~al.(2025)Li, Cheng, Feng, Liu, and Wang]{li2025mask}
Yongkang Li, Tianheng Cheng, Bin Feng, Wenyu Liu, and Xinggang Wang.
\newblock Mask-adapter: The devil is in the masks for open-vocabulary segmentation.
\newblock In \emph{Proceedings of the Computer Vision and Pattern Recognition Conference}, pages 14998--15008, 2025.

\bibitem[Liang et~al.(2023{\natexlab{a}})Liang, Wu, Dai, Li, Zhao, Zhang, Zhang, Vajda, and Marculescu]{liang2023open}
Feng Liang, Bichen Wu, Xiaoliang Dai, Kunpeng Li, Yinan Zhao, Hang Zhang, Peizhao Zhang, Peter Vajda, and Diana Marculescu.
\newblock Open-vocabulary semantic segmentation with mask-adapted clip.
\newblock In \emph{Proceedings of the IEEE/CVF conference on computer vision and pattern recognition}, pages 7061--7070, 2023{\natexlab{a}}.

\bibitem[Liang et~al.(2023{\natexlab{b}})Liang, Wu, Dai, Li, Zhao, Zhang, Zhang, Vajda, and Marculescu]{liang23cvpr}
Feng Liang, Bichen Wu, Xiaoliang Dai, Kunpeng Li, Yinan Zhao, Hang Zhang, Peizhao Zhang, Peter Vajda, and Diana Marculescu.
\newblock Open-vocabulary semantic segmentation with mask-adapted clip.
\newblock In \emph{Proceedings of the IEEE/CVF conference on computer vision and pattern recognition}, pages 7061--7070, 2023{\natexlab{b}}.

\bibitem[Lin et~al.(2014)Lin, Maire, Belongie, Hays, Perona, Ramanan, Doll{\'a}r, and Zitnick]{lin14eccv}
Tsung-Yi Lin, Michael Maire, Serge Belongie, James Hays, Pietro Perona, Deva Ramanan, Piotr Doll{\'a}r, and C~Lawrence Zitnick.
\newblock Microsoft coco: Common objects in context.
\newblock In \emph{European Conference on Computer Vision (ECCV)}, pages 740--755. Springer, 2014.

\bibitem[Liu et~al.(2022)Liu, Mao, Wu, Feichtenhofer, Darrell, and Xie]{liu2022convnet}
Zhuang Liu, Hanzi Mao, Chao-Yuan Wu, Christoph Feichtenhofer, Trevor Darrell, and Saining Xie.
\newblock A convnet for the 2020s.
\newblock In \emph{Proceedings of the IEEE/CVF conference on computer vision and pattern recognition}, pages 11976--11986, 2022.

\bibitem[Long et~al.(2015)Long, Shelhamer, and Darrell]{long15cvpr}
Jonathan Long, Evan Shelhamer, and Trevor Darrell.
\newblock Fully convolutional networks for semantic segmentation.
\newblock In \emph{Proceedings of the IEEE conference on computer vision and pattern recognition}, pages 3431--3440, 2015.

\bibitem[Martinovi{\'c} et~al.(2025)Martinovi{\'c}, {\v{S}}ari{\'c}, Or{\v{s}}i{\'c}, Kristan, and {\v{S}}egvi{\'c}]{martinovic2025dearli}
Ivan Martinovi{\'c}, Josip {\v{S}}ari{\'c}, Marin Or{\v{s}}i{\'c}, Matej Kristan, and Sini{\v{s}}a {\v{S}}egvi{\'c}.
\newblock Dearli: Decoupled enhancement of recognition and localization for semi-supervised panoptic segmentation.
\newblock \emph{arXiv preprint arXiv:2507.10118}, 2025.

\bibitem[Mukhoti et~al.(2023)Mukhoti, Lin, Poursaeed, Wang, Shah, Torr, and Lim]{mukhoti2023open}
Jishnu Mukhoti, Tsung-Yu Lin, Omid Poursaeed, Rui Wang, Ashish Shah, Philip~HS Torr, and Ser-Nam Lim.
\newblock Open vocabulary semantic segmentation with patch aligned contrastive learning.
\newblock In \emph{Proceedings of the IEEE/CVF Conference on Computer Vision and Pattern Recognition}, pages 19413--19423, 2023.

\bibitem[Naeem et~al.(2024)Naeem, Xian, Zhai, Hoyer, Van~Gool, and Tombari]{naeem2024silc}
Muhammad~Ferjad Naeem, Yongqin Xian, Xiaohua Zhai, Lukas Hoyer, Luc Van~Gool, and Federico Tombari.
\newblock Silc: Improving vision language pretraining with self-distillation.
\newblock In \emph{European Conference on Computer Vision}, pages 38--55. Springer, 2024.

\bibitem[Neuhold et~al.(2017)Neuhold, Ollmann, Rota~Bulo, and Kontschieder]{neuhold2017mapillary}
Gerhard Neuhold, Tobias Ollmann, Samuel Rota~Bulo, and Peter Kontschieder.
\newblock The mapillary vistas dataset for semantic understanding of street scenes.
\newblock In \emph{Proceedings of the IEEE international conference on computer vision}, pages 4990--4999, 2017.

\bibitem[Radford et~al.(2021)Radford, Kim, Hallacy, Ramesh, Goh, Agarwal, Sastry, Askell, Mishkin, Clark, et~al.]{radford21icml}
Alec Radford, Jong~Wook Kim, Chris Hallacy, Aditya Ramesh, Gabriel Goh, Sandhini Agarwal, Girish Sastry, Amanda Askell, Pamela Mishkin, Jack Clark, et~al.
\newblock Learning transferable visual models from natural language supervision.
\newblock In \emph{International conference on machine learning}, pages 8748--8763. PmLR, 2021.

\bibitem[Ren et~al.(2015)Ren, He, Girshick, and Sun]{ren15neurips}
Shaoqing Ren, Kaiming He, Ross Girshick, and Jian Sun.
\newblock Faster r-cnn: Towards real-time object detection with region proposal networks.
\newblock \emph{Advances in neural information processing systems}, 28, 2015.

\bibitem[Ronneberger et~al.(2015)Ronneberger, Fischer, and Brox]{ronneberger15miccai}
Olaf Ronneberger, Philipp Fischer, and Thomas Brox.
\newblock U-net: Convolutional networks for biomedical image segmentation.
\newblock In \emph{International Conference on Medical image computing and computer-assisted intervention (MICCAI)}, pages 234--241. Springer, 2015.

\bibitem[Schuhmann et~al.(2022)Schuhmann, Beaumont, Vencu, Gordon, Wightman, Cherti, Coombes, Katta, Mullis, Wortsman, et~al.]{schuhmann22neurips}
Christoph Schuhmann, Romain Beaumont, Richard Vencu, Cade Gordon, Ross Wightman, Mehdi Cherti, Theo Coombes, Aarush Katta, Clayton Mullis, Mitchell Wortsman, et~al.
\newblock Laion-5b: An open large-scale dataset for training next generation image-text models.
\newblock \emph{Advances in Neural Information Processing Systems}, 35:\penalty0 25278--25294, 2022.

\bibitem[Sun et~al.(2023)Sun, Fang, Wu, Wang, and Cao]{sun2023eva}
Quan Sun, Yuxin Fang, Ledell Wu, Xinlong Wang, and Yue Cao.
\newblock Eva-clip: Improved training techniques for clip at scale.
\newblock \emph{arXiv preprint arXiv:2303.15389}, 2023.

\bibitem[Tschannen et~al.(2025)Tschannen, Gritsenko, Wang, Naeem, Alabdulmohsin, Parthasarathy, Evans, Beyer, Xia, Mustafa, et~al.]{tschannen2025siglip}
Michael Tschannen, Alexey Gritsenko, Xiao Wang, Muhammad~Ferjad Naeem, Ibrahim Alabdulmohsin, Nikhil Parthasarathy, Talfan Evans, Lucas Beyer, Ye Xia, Basil Mustafa, et~al.
\newblock Siglip 2: Multilingual vision-language encoders with improved semantic understanding, localization, and dense features.
\newblock \emph{arXiv preprint arXiv:2502.14786}, 2025.

\bibitem[Wang et~al.(2024)Wang, Mei, and Yuille]{wang2024sclip}
Feng Wang, Jieru Mei, and Alan Yuille.
\newblock Sclip: Rethinking self-attention for dense vision-language inference.
\newblock In \emph{European Conference on Computer Vision}, pages 315--332. Springer, 2024.

\bibitem[Wang et~al.(2021)Wang, Zhu, Adam, Yuille, and Chen]{wang21cvpr}
Huiyu Wang, Yukun Zhu, Hartwig Adam, Alan Yuille, and Liang-Chieh Chen.
\newblock Max-deeplab: End-to-end panoptic segmentation with mask transformers.
\newblock In \emph{Proceedings of the IEEE/CVF conference on computer vision and pattern recognition}, pages 5463--5474, 2021.

\bibitem[Wang et~al.(2023)Wang, Dai, Chen, Huang, Li, Zhu, Hu, Lu, Lu, Li, et~al.]{wang23cvpr}
Wenhai Wang, Jifeng Dai, Zhe Chen, Zhenhang Huang, Zhiqi Li, Xizhou Zhu, Xiaowei Hu, Tong Lu, Lewei Lu, Hongsheng Li, et~al.
\newblock Internimage: Exploring large-scale vision foundation models with deformable convolutions.
\newblock In \emph{Proceedings of the IEEE/CVF conference on computer vision and pattern recognition}, pages 14408--14419, 2023.

\bibitem[Wu et~al.(2023)Wu, Zhang, Xu, Jin, Li, Liu, and Loy]{wu2023clipself}
Size Wu, Wenwei Zhang, Lumin Xu, Sheng Jin, Xiangtai Li, Wentao Liu, and Chen~Change Loy.
\newblock Clipself: Vision transformer distills itself for open-vocabulary dense prediction.
\newblock \emph{arXiv preprint arXiv:2310.01403}, 2023.

\bibitem[Xie et~al.(2024)Xie, Cao, Xie, Khan, and Pang]{xie2024cvpr}
Bin Xie, Jiale Cao, Jin Xie, Fahad~Shahbaz Khan, and Yanwei Pang.
\newblock Sed: A simple encoder-decoder for open-vocabulary semantic segmentation.
\newblock In \emph{Proceedings of the IEEE/CVF Conference on Computer Vision and Pattern Recognition (CVPR)}, pages 3426--3436, 2024.

\bibitem[Xu et~al.(2023{\natexlab{a}})Xu, Xie, Tan, Huang, Howes, Sharma, Li, Ghosh, Zettlemoyer, and Feichtenhofer]{xu2023demystifying}
Hu Xu, Saining Xie, Xiaoqing~Ellen Tan, Po-Yao Huang, Russell Howes, Vasu Sharma, Shang-Wen Li, Gargi Ghosh, Luke Zettlemoyer, and Christoph Feichtenhofer.
\newblock Demystifying clip data.
\newblock \emph{arXiv preprint arXiv:2309.16671}, 2023{\natexlab{a}}.

\bibitem[Xu et~al.(2022)Xu, De~Mello, Liu, Byeon, Breuel, Kautz, and Wang]{xu2022groupvit}
Jiarui Xu, Shalini De~Mello, Sifei Liu, Wonmin Byeon, Thomas Breuel, Jan Kautz, and Xiaolong Wang.
\newblock Groupvit: Semantic segmentation emerges from text supervision.
\newblock In \emph{Proceedings of the IEEE/CVF conference on computer vision and pattern recognition}, pages 18134--18144, 2022.

\bibitem[Xu et~al.(2023{\natexlab{b}})Xu, Hou, Zhang, Feng, Wang, Qiao, and Xie]{xu2023cvprlearningfromnls}
Jilan Xu, Junlin Hou, Yuejie Zhang, Rui Feng, Yi Wang, Yu Qiao, and Weidi Xie.
\newblock Learning open-vocabulary semantic segmentation models from natural language supervision.
\newblock In \emph{Proceedings of the IEEE/CVF Conference on Computer Vision and Pattern Recognition (CVPR)}, pages 2935--2944, 2023{\natexlab{b}}.

\bibitem[Xu et~al.(2023{\natexlab{c}})Xu, Liu, Vahdat, Byeon, Wang, and De~Mello]{xu2023open}
Jiarui Xu, Sifei Liu, Arash Vahdat, Wonmin Byeon, Xiaolong Wang, and Shalini De~Mello.
\newblock Open-vocabulary panoptic segmentation with text-to-image diffusion models.
\newblock In \emph{Proceedings of the IEEE/CVF Conference on Computer Vision and Pattern Recognition}, pages 2955--2966, 2023{\natexlab{c}}.

\bibitem[Xu et~al.(2023{\natexlab{d}})Xu, Zhang, Wei, Hu, and Bai]{xu2023side}
Mengde Xu, Zheng Zhang, Fangyun Wei, Han Hu, and Xiang Bai.
\newblock Side adapter network for open-vocabulary semantic segmentation.
\newblock In \emph{Proceedings of the IEEE/CVF conference on computer vision and pattern recognition}, pages 2945--2954, 2023{\natexlab{d}}.

\bibitem[Yu et~al.(2022)Yu, Wang, Qiao, Collins, Zhu, Adam, Yuille, and Chen]{yu22eccv}
Qihang Yu, Huiyu Wang, Siyuan Qiao, Maxwell Collins, Yukun Zhu, Hartwig Adam, Alan Yuille, and Liang-Chieh Chen.
\newblock k-means mask transformer.
\newblock In \emph{European Conference on Computer Vision}, pages 288--307. Springer, 2022.

\bibitem[Yu et~al.(2023)Yu, He, Deng, Shen, and Chen]{yu23neurips}
Qihang Yu, Ju He, Xueqing Deng, Xiaohui Shen, and Liang-Chieh Chen.
\newblock Convolutions die hard: Open-vocabulary segmentation with single frozen convolutional clip.
\newblock \emph{Advances in Neural Information Processing Systems}, 36:\penalty0 32215--32234, 2023.

\bibitem[Zhai et~al.(2022)Zhai, Wang, Mustafa, Steiner, Keysers, Kolesnikov, and Beyer]{zhai2022lit}
Xiaohua Zhai, Xiao Wang, Basil Mustafa, Andreas Steiner, Daniel Keysers, Alexander Kolesnikov, and Lucas Beyer.
\newblock Lit: Zero-shot transfer with locked-image text tuning.
\newblock In \emph{Proceedings of the IEEE/CVF conference on computer vision and pattern recognition}, pages 18123--18133, 2022.

\bibitem[Zhai et~al.(2023)Zhai, Mustafa, Kolesnikov, and Beyer]{zhai2023sigmoid}
Xiaohua Zhai, Basil Mustafa, Alexander Kolesnikov, and Lucas Beyer.
\newblock Sigmoid loss for language image pre-training.
\newblock In \emph{Proceedings of the IEEE/CVF international conference on computer vision}, pages 11975--11986, 2023.

\bibitem[Zhao et~al.(2017)Zhao, Shi, Qi, Wang, and Jia]{zhao17cvpr}
Hengshuang Zhao, Jianping Shi, Xiaojuan Qi, Xiaogang Wang, and Jiaya Jia.
\newblock Pyramid scene parsing network.
\newblock In \emph{Proceedings of the IEEE conference on computer vision and pattern recognition}, pages 2881--2890, 2017.

\bibitem[Zhou et~al.(2017)Zhou, Zhao, Puig, Fidler, Barriuso, and Torralba]{zhou17cvpr}
Bolei Zhou, Hang Zhao, Xavier Puig, Sanja Fidler, Adela Barriuso, and Antonio Torralba.
\newblock Scene parsing through ade20k dataset.
\newblock In \emph{Proceedings of the IEEE Conference on Computer Vision and Pattern Recognition (CVPR)}, pages 5122--5130, 2017.

\bibitem[Zhou et~al.(2022)Zhou, Loy, and Dai]{zhou22eccv}
Chong Zhou, Chen~Change Loy, and Bo Dai.
\newblock Extract free dense labels from clip.
\newblock In \emph{European Conference on Computer Vision}, pages 696--712. Springer, 2022.

\end{thebibliography}
}

\end{document}